\documentclass[10pt, conference, compsocconf]{IEEEtran}
\pdfoutput=1
\usepackage{hyperref}
\hypersetup{
	colorlinks=true,
	linkcolor=black,
	citecolor=black,
	urlcolor=blue,
}
\usepackage[caption=false,font=normalsize,labelfont=sf,textfont=sf]{subfig}
\usepackage{algorithm, algorithmic}
\usepackage[noadjust]{cite}
\usepackage{fancyvrb}
\usepackage{amsmath, amsfonts}
\usepackage{soul}
\usepackage{graphicx}
\usepackage{color}
\graphicspath{{figures/}}

\newlength{\HalfPage}
\newcommand{\etal}{{et al.}}
\newcommand{\selfstar}{{\em self-}$\ast$}

\newcommand{\CalloutTable}[1]{Table~\ref{#1}}
\newcommand{\CalloutFigure}[1]{Figure~\ref{#1}}

\newcommand{\CalloutSection}[1]{Section~\ref{#1}}

\begin{document}
\sloppy

\title{Leveraging Evolutionary Search to Discover \\Self-Adaptive and Self-Organizing Cellular Automata}

\author{\IEEEauthorblockN{
David B.~Knoester\IEEEauthorrefmark{1},
Heather J.~Goldsby\IEEEauthorrefmark{2}, and
Christoph C.~Adami\IEEEauthorrefmark{1}}
\IEEEauthorblockA{\IEEEauthorrefmark{1}Dept.~of Microbiology and Molecular Genetics\\
Michigan State University\\
East Lansing, Michigan 48824\\
Email: \{dk,adami\}@msu.edu}
\IEEEauthorblockA{\IEEEauthorrefmark{2}Dept. of Biology\\
University of Washington\\
Seattle, Washington 98195\\
Email: goldsby@uw.edu}}

\maketitle

\begin{abstract}
Building self-adaptive and self-organizing (SASO) systems is a challenging problem, in part because SASO principles are not yet well understood and few platforms exist for exploring them.  \emph{Cellular automata} (CA) are a well-studied approach to exploring the principles underlying self-organization.  A CA comprises a lattice of cells whose states change over time based on a discrete update function.  One challenge to developing CA is that the relationship of an update function, which describes the local behavior of each cell, to the global behavior of the entire CA is often unclear.  As a result, many researchers have used stochastic search techniques, such as evolutionary algorithms, to automatically discover update functions that produce a desired global behavior.  However, these update functions are typically defined in a way that does not provide for self-adaptation.  Here we describe an approach to discovering CA update functions that are both self-adaptive and self-organizing.  Specifically, we use a novel evolutionary algorithm-based approach to discover finite state machines (FSMs) that implement update functions for CA.  We show how this approach is able to evolve FSM-based update functions that perform well on the density classification task for 1-, 2-, and 3-dimensional CA.  Moreover, we show that these FSMs are self-adaptive, self-organizing, and highly scalable, often performing well on CA that are orders of magnitude larger than those used to evaluate performance during the evolutionary search.  These results demonstrate that CA are a viable platform for studying the integration of self-adaptation and self-organization, and strengthen the case for using evolutionary algorithms as a component of SASO systems.
\end{abstract}


\section{Introduction}\label{s:intro}

Engineering large-scale distributed computing systems that are resilient in the face of dynamic and often hostile environments is a challenging problem.  {\em Cyber-physical} systems~\cite{Wolf:2009uh}, where computing systems interact with the physical world, further exacerbate these difficulties.  {\em Autonomic computing}, so-named for its relation to the autonomic nervous system~\cite{Kephart:2003uq}, was envisioned as a means to address these challenges.  Frameworks for building such systems are characterized by fault-tolerance, self-healing, self-organization, and self-adaptivity.  Collectively, these are known as the {\selfstar} properties~\cite{Babaoglu:2005wb}.

While progress has been made on building computational {\selfstar} systems, nature provides us with many examples of systems that already exhibit some, if not all, of these properties.  For example, self-organizing systems can be found in the formation of inorganic crystals and the complex social behaviors of honeybees, ants, and termites~\cite{Holldobler:2008ws}.  Many organisms also exhibit a form of self-adaptation known as {\em plasticity}, where their behavior and/or phenotype may change in response to environmental stimuli~\cite{Bradshaw:1965uh,WestEberhard:1989wq,Agrawal:2001ji}.  For example, plants may have different flowering times, stem characteristics, or leaf shape based on the resources present in their environment~\cite{Bradshaw:1965uh}.  While many researchers have taken inspiration from biology to develop {\selfstar} systems, it remains challenging to develop a single computational framework that is {\em both} self-adaptive and self-organizing (SASO).
\footnote{We note that there is some overloading of the term {\em adaptation}.  In biology, {\em adaptation} refers to how a population changes in a Darwinian manner as a response to selection pressures acting upon it.  In evolutionary computation, {\em self-adaptation} refers to changing parameters (e.g., mutation rate) of an evolutionary algorithm in the midst of a search~\cite{Kramer:2010fg}.  In computing systems, {\em adaptive software} generally refers to software that is able to exhibit a behavioral change at runtime.  Here, we use self-adaptation in the computing systems sense.}

\begin{figure}
\centering\includegraphics[width=.4\textwidth]{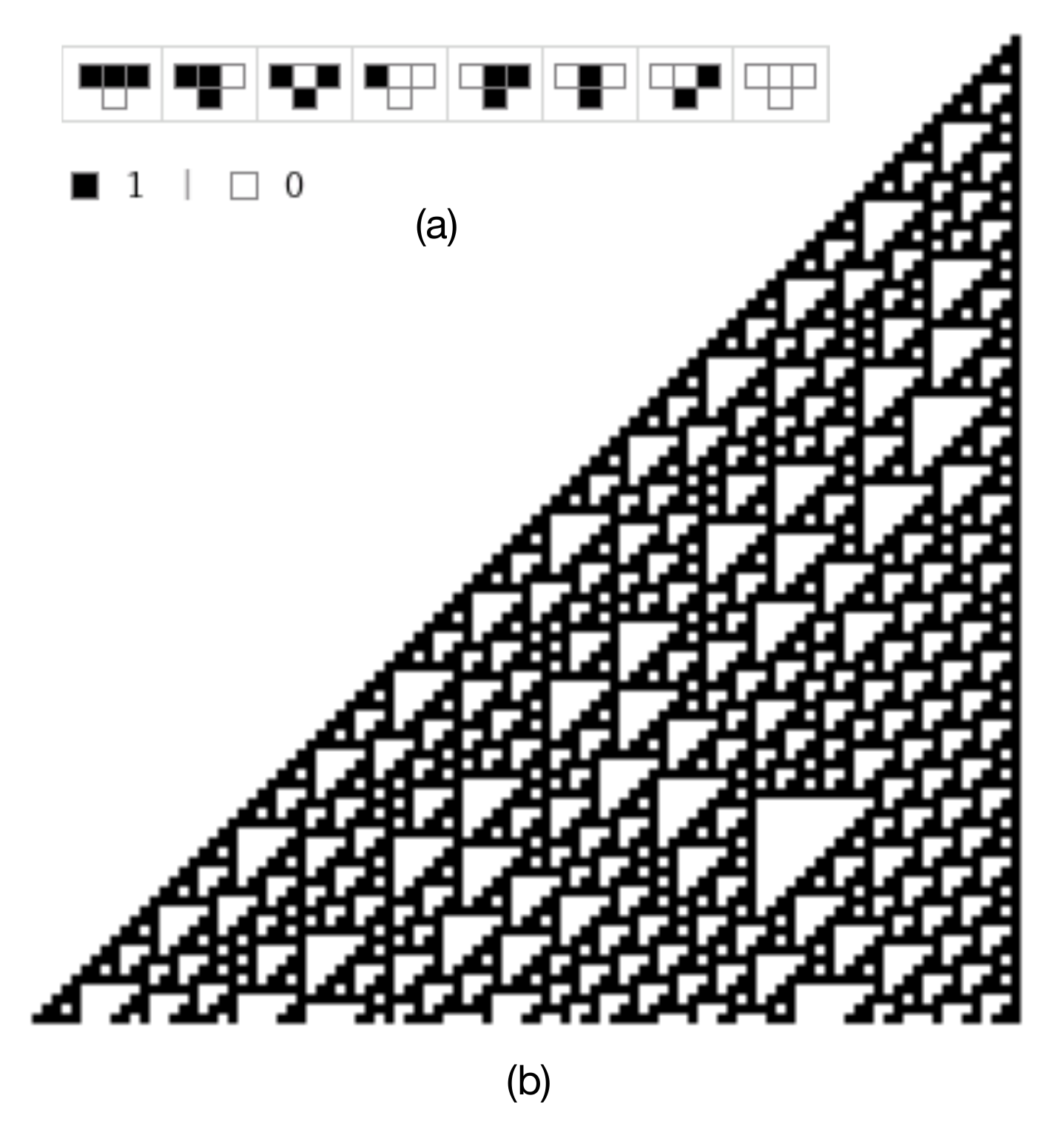}
\caption{An elementary CA known as ``Rule 110''~\cite{Wolfram:1994tc}, which has been shown to be Turing complete~\cite{Cook:2004tt}.  (a) The update function for Rule 110.  Each possible combination of states for the neighborhood of a cell is given in the top row, while the output (next) state for that cell is given in the bottom row.  This update function implements the formula: $(q \wedge \neg p) \vee (q \oplus r)$.  (b) CA behavior for one initial configuration when updated via Rule 110.  Figures from Wolfram$|$Alpha.}
\label{f:ca-example}
\end{figure}

{\em Cellular automata} (CA) are a form of discrete dynamical system where a lattice of $N$ stateful {\em cells} change over time.  
CA were popularized in Conway's ``Game of Life''~\cite{Gardner:1970to} and rest on a foundation of research established by von Neumann, Burks, and Holland~\cite{Neumann:1966wc,Holland:1970fy}, later expanded by Wolfram~\cite{Wolfram:1994tc}, and have been used to study a variety of phenomena.  Cells in a typical CA are always in exactly one of a finite number of states, $s$.  In most CA (and also here), cell states are binary: $s=\{0,1\}$.  The state of all cells in a CA is known as a {\em configuration}.  The initial configuration (time $t=0$) of a CA is typically random.  Configurations change over time based on a discrete {\em update function}, $\Phi(n_i)$, which computes the state of cell $c_i$ at time $t+1$ based on the states of the cells in its neighborhood $n_i$ at time $t$.  Neighborhoods are defined by the topology of the underlying lattice of cells and a {\em radius} $r$, which typically varies from 1 to 3.  For example, \CalloutFigure{f:ca-example} depicts ``Rule 110,'' a well-known 1-dimensional (1D) with $r=1$.  Here, $n_i$ consists of the states of three cells: $\{c_{i-1}, c_i, c_{i+1}\}$ (by convention, $n_i$ includes the state of cell $c_i$).  We note that the boundary conditions of most CA are periodic, i.e., neighborhoods ``wrap around'' such that a 1D CA is a ring, a 2D CA is a torus, etc.

While update functions for CA often result in intriguing patterns, the relationship of an update function to global behavior is often unclear.  Indeed, Land and Belew have proved that perfect solutions to the canonical density classification problem, where a CA must calculate the most common state in its initial configuration, do not exist~\cite{Land:1995ha} (this task will be described in more detail in \CalloutSection{s:methods}).  As a result, many have turned to stochastic search techniques, such as evolutionary algorithms, to generate appropriate update functions for a desired global behavior~\cite{Darabos:2013if,Mitchell:1996ut,Andre:1996wa,Breukelaar:2005jz,Chavoya:2006uz}.  For example, Mitchell~{\etal} have shown that genetic algorithms are capable of discovering update functions for 1D CA with $N=149$~\cite{Mitchell:1996ut}, while Andre~{\etal} demonstrated a genetic programming approach that discovered high-performing update functions for solving the density classification problem~\cite{Andre:1996wa}.   More recently, Darabos~{\etal} have used evolutionary algorithms to explore the relationship of the update function to the topology of the neighborhood surrounding each cell~\cite{Darabos:2013if}.  While these studies advance our understanding of the principles of self-organization, they do not include self-adaptive elements.

Here, we describe an approach for studying self-adaptive and self-organizing systems using CA. Specifically, we present a novel evolutionary algorithm that discovers finite state machines (FSMs) that implement CA update functions and allow both self-organization and self-adaptation.  We demonstrate these principles by evolving FSM-specified update functions that solve the density classification task in 1-, 2-, and 3-dimensions.  We then show that the evolved FSMs are self-organizing and highly scalable, often performing well on CA that are orders of magnitude larger than those used during evolution.  Finally, we describe how the evolved FSMs are self-adaptive, in that their behavior is dependent upon internal states that are effectively a simple form of evolved memory.  These results support the use of CA as a study system for exploring SASO principles, illustrate one method by which self-adaptation can be integrated into self-organizing systems, and further strengthen the case for the use of evolutionary algorithms as a component of SASO systems.

\section{Related Work}\label{s:related}

From von Neumann's early work on self-reproducing automata~\cite{Neumann:1966wc} to Wolfram's later work on complexity~\cite{Wolfram:1994tc}, CA have been used to study self-organizing systems across a broad range of domains.  For example, CA have been used in image processing~\cite{Rosin:2006fd}, to simulate distributed behavior of flocks of birds and schools of fish~\cite{Reynolds:1987iz}, as model systems for artificial life and parallel processing~\cite{Langton:1986ja,Bandini:2001ch,Sayama:2004bf}, and for calculating convex hulls and Voronoi diagrams~\cite{Maignan:2011cd}.  Traditionally, CA update functions are specified as a binary lookup table and rely exclusively on the current state of a cell and its neighbors. However, some CA also have the capacity for {\em memory}, where the update function takes into account the previous state(s) of a cell and the cells in its neighborhood~\cite{AlonsoSanz:2012ub}. Other CA replace the typical update function with discrete- or continuous- time artificial neurons, and are known as {\em cellular neural networks}~\cite{Chua:2002vl,Harrer:1992it}. Here, to enable self-adaptation, we use an evolutionary algorithm to discover {\em both} state transition logic and memory characteristics that define a CA update function.

Evolutionary algorithms (EAs) are search techniques inspired by the process of evolution by natural selection~\cite{Eiben:2003tf}. EAs have a long history of being used to solve optimization problems in engineering~\cite{Fogel:1994eg}, and as a tool to study and develop self-organizing systems~\cite{Knoester:2008wk,Knoester:2009vl,Knoester:2011gp,Beckmann:2007tt,Goldsby:2014df}. A traditional EA comprises a fixed-size population of candidate solutions to a given problem. Each generation, candidate solutions within a population are evaluated to establish how well they solve the problem. Subsequently, a subset of individuals are mutated and recombined to produce new solutions. The next generation of the population is then constructed by selecting individuals from the current generation and newly-produced offspring solutions.  Each candidate solution has a genome, which describes its solution to the problem. This genome can be quite simple (a bitstring), or more complex (a tree of Lisp-like $s$-expressions).  EAs have previously been applied to the discovery of simple update functions for CA, most frequently using a genetic algorithm to define a binary rule table~\cite{Mitchell:1996ut,Breukelaar:2005jz,Chavoya:2006uz}. However, there have been approaches that also define more complex update functions and neighborhood topologies~\cite{Andre:1996wa,Darabos:2013if}.  Historically, density classification on 1D CA with $N=149$ has been recognized as a demanding problem for genetic algorithms, and a variety of approaches have been evaluated~\cite{Stone:2009ki}.  The work presented here differs in that (a) we present results for 1-, 2-, and 3D CA and (b) the evolutionary algorithm used here is {\em allowed} to evolve self-adaptation capabilities, but it is by no means certain that it will do so.

{\em Evolutionary programming} is a kind of EA that focuses on the discovery of finite state machines~\cite{Fogel:1966wb}.  Evolutionary programming has frequently been used to address questions in machine intelligence~\cite{Fogel:2006uy,Fogel:2008hy,Fogel:1999vs,Fogel:1991va}. In contrast, the approach taken in this paper is inspired by the evolution of {\em Markov networks}, which are networks of random variables with the Markov property, such that edges between nodes encode arbitrary fuzzy logic gates~\cite{Edlund:2011kt}.  Markov networks have been shown to be remarkably effective at exploring questions related to the dynamics of collective behavior, including predator confusion~\cite{Olson:2013kx} and the formation of the selfish herd~\cite{Olson:2013ko}.  While we do not make use of stochastic behavior in this paper, Fat\'es has shown that stochastic approaches to the density classification task can perform quite well~\cite{Fates:2013kp}, a subject which we hope to explore in future work.


\section{Methods}\label{s:methods}

\subsection{Density classification task}
\begin{figure}
\centering
\includegraphics[width=\HalfPage]{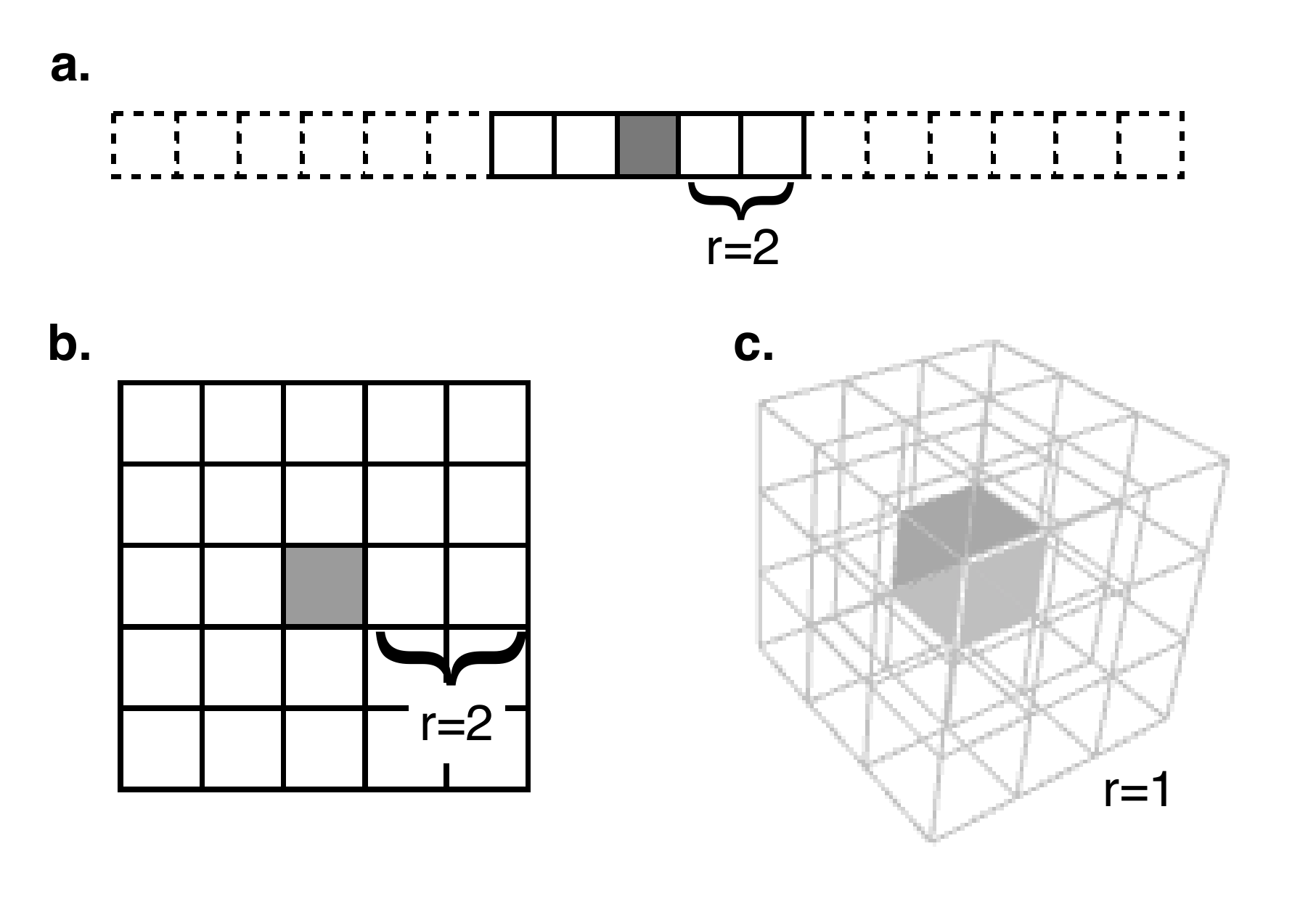}
\caption{(a) 1D cellular automata (CA) with radius $r=2$.  Each cell in the array determines its state at time $t+1$ based on the states of the 5 cells in its neighborhood (itself plus the state of its $2*r$ neighbors).  (b) 2D CA with radius $r=2$; each cell determines its state at time $t+1$ based on the states of 25 cells.  (c)  3D CA with radius $r=1$; each cell determines its state at time $t+1$ based on the states of 27 cells.  2D and 3D topologies both use a Moore neighborhood.}
\label{f:ca-dimensions}
\end{figure}

Density classification is a challenging and yet tractable distributed computation task for exploring emergent behavior within CA~\cite{Mitchell:1996ut}.  This task requires the CA to sense the density of cells in 0 (or 1) in its initial configuration and modify all cell states accordingly.  Specifically, given an initial (random) configuration of cell states $IC$, where the possible states $s=\{0,1\}$, the desired global behavior of the CA is to determine whether a threshold density $\rho_c$ of cells in $IC$ are 0 or 1.  Here (as in~\cite{Mitchell:1996ut}) we use $\rho_c=1/2$.  For example, if $\rho_0$ is the density of 1s in a given initial configuration, then the CA should reach a state of all 1s within $M$ time steps if  $\rho_0 > \rho_c$, otherwise it should reach a state of all 0s within $M$ time steps.  $M$ is a parameter that depends on the size of the lattice, and is set to 2 times the number of cells.  In this paper, we evolve the update rules for CA that address the density classification task in three different dimensions and topologies (depicted in \CalloutFigure{f:ca-dimensions}).

\subsection{Evolving update functions}\label{s:fitness-function}
Rather than describing the behavior of cells using a binary rule table, here we use an evolutionary algorithm to discover finite state machines that implement update functions for the density classification task.  The candidate solutions (FSMs) in this EA are encoded in evolving genomes (described below).  As part of the evaluation process, we test each FSM on a set of randomly generated initial conditions ($IC$s).  Specifically, during each evaluation of an $IC$, all cells in the CA independently execute their own copy of the genetically encoded FSM.  For example, in a 1D CA of 35 cells, there are 35 identical copies of an FSM that has been translated from a genome.  During evaluation, the FSMs corresponding to these 35 cells each take input from their neighborhood and produce a single output bit, which is the state of their corresponding cell at time $t+1$.

As in~\cite{Mitchell:1996ut}, the initial conditions are generated based on a density that is drawn from a uniform distribution $[0,1]$. This density determines the number of 1s present in the $IC$ (the positions of these 1s are randomly shuffled; all other bits are 0).  Finally, we draw ICs from three different sets: those where the density is strictly less than 0.5, those with density greater than 0.5, and a binomial distribution with $p=0.5$.  This latter case is considered the most difficult, as on average there will be an equal number of 0s and 1s in the $IC$s. 

The {\em fitness} of a given genome, which is used by the EA to calculate survival and reproduction probabilities, is calculated as follows.  First, we translate the genome into an FSM (this process is described in more detail below).  This FSM is then evaluated on 100 different initial conditions, and the mean number of correct classifications is calculated.  No credit is given for partial solutions.  To reduce the chance of an FSM being ``lucky'' during fitness evaluation, we change the set of initial conditions that are used to evaluate an FSM every generation and average the fitness of a genome over the fitness of its immediate 10 ancestors.  This approach is similar to noise mitigation via multiple fitness evaluations~\cite{Jin:2005cm}, but avoids the additional computational overhead.

\subsection{Genetic encoding of finite state machines}
\begin{figure}
\centering
\includegraphics[width=.45\textwidth]{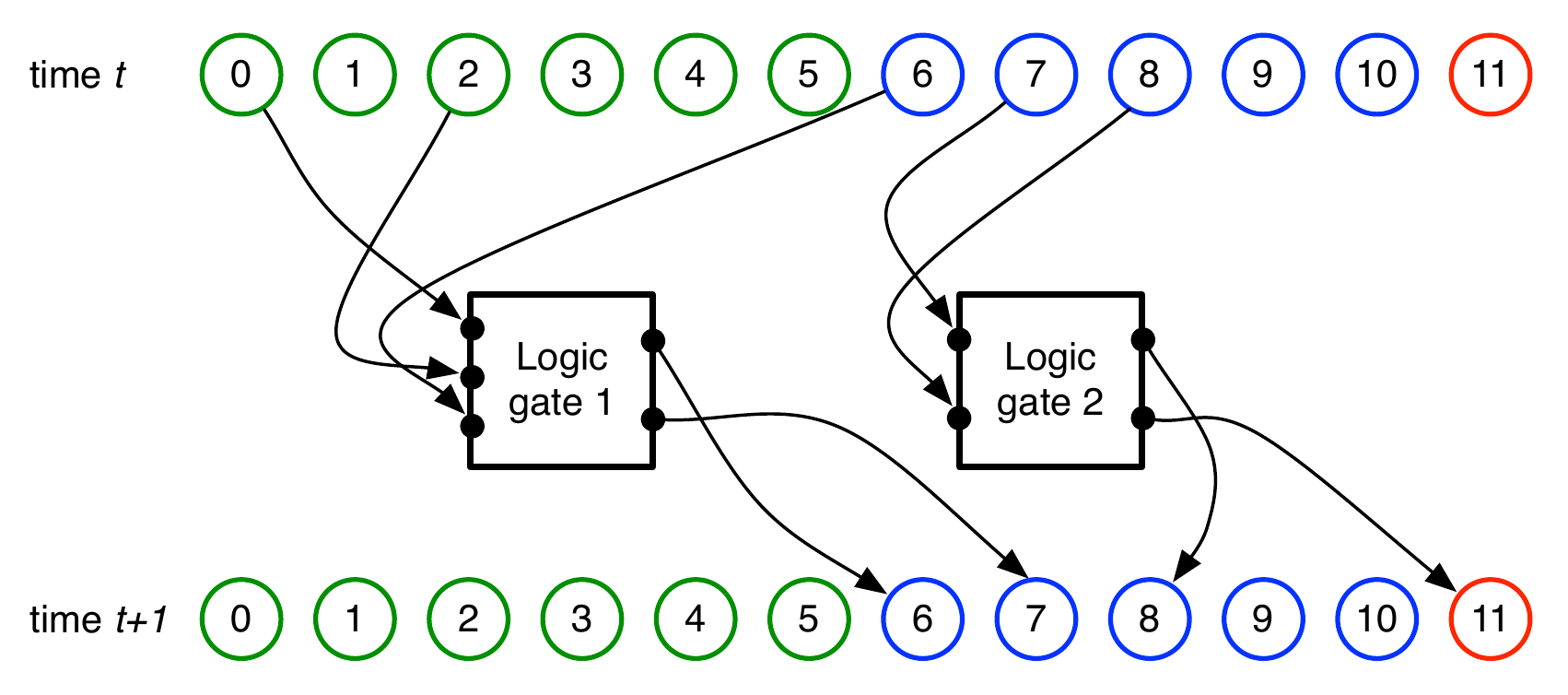}
\caption{A simple evolved finite state machine.  The inputs and outputs for two logic gates are drawn from the predefined global state variables.  Here, gate 1 takes 3 inputs at time $t$: 2 from input state variables (green) and 1 from a hidden state variable.  Gate 1 also writes into 2 hidden state variables (blue) at time $t+1$.  Similarly, gate 2 takes input from 2 hidden states, and writes to both an output state (red) and a hidden state.}
\label{f:fsm-example}
\end{figure}

We base our approach on the evolution of Markov networks~\cite{Edlund:2011kt} and adapt it to evolve FSMs.  As shown in \CalloutFigure{f:fsm-example}, we begin by defining a fixed number of binary state variables.  Each state variable is either an input that can accept state input from a given cell's neighborhood; an output that produces the next state of a cell; or a hidden state variable that is internal to a cell's FSM and can be used to implement memory.  Updates to these state variables are controlled via evolved logic gates that are translated from the genome.

\begin{figure}
\centering
\includegraphics[width=.45\textwidth]{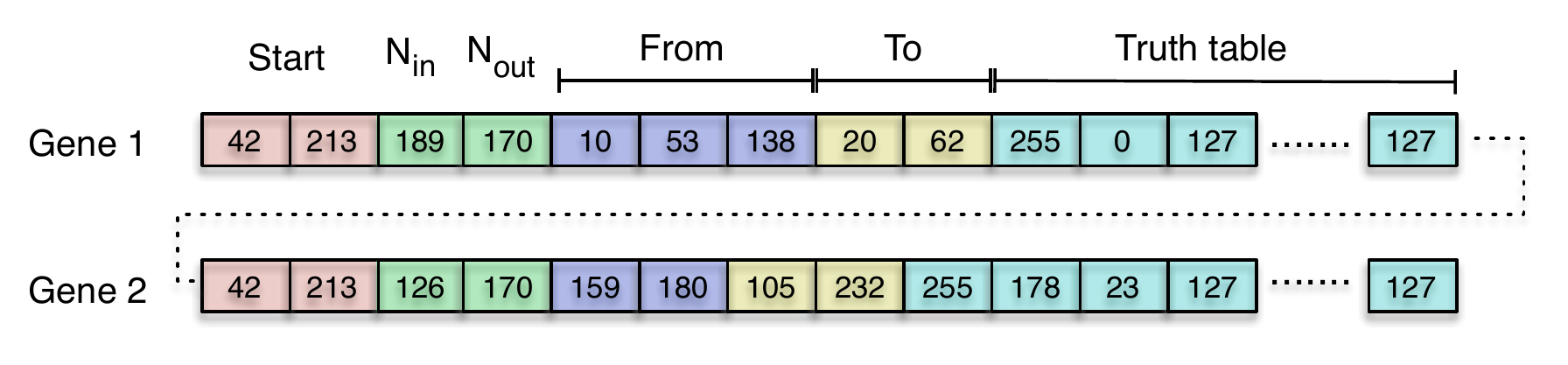}
\caption{Example genome with two genes that encode the two gates in Fig.~\ref{f:fsm-example}. The doublet (42, 213) is the start codon that marks the beginning of a gene (red blocks). The next two codons describe the fan-in and fan-out of the gate (green blocks), and the following codons map the inputs (blue blocks) and outputs (yellow blocks) of that gate to state variables.  The remaining codons encode the gate's truth table (cyan blocks).  Numeric values in the genome are converted via a mod operation into the appropriate range.}
\label{f:fsm-genome}
\end{figure}

Figure~\CalloutFigure{f:fsm-genome} depicts the genome encoding for an FSM.  The logic gates that comprise an FSM are embedded in a circular list of integers (the genome) that contains a series of {\em genes}.  Each gene encodes a single logic gate, and defines that gate's fan-in and fan-out (number of inputs and outputs), underlying truth table that defines the logic operation performed by the gate, and the mapping of inputs and outputs to this gate to the global input, output, and hidden state variables.  These logic gates operate much like a neuron in an artificial neural network, where all input and output from gates in the FSM is via these global state variables (it may be helpful to think of these state variables as a scoreboard, where each gate synchronously updates a global value).  The beginning of a gene is indicated by a {\em start codon} (a specified sequence of two integers) and the gene's size is determined by the size of the corresponding gate's truth table.  The truth table for a given logic gate is defined by the $2^{n_{in}}$ integers following the output state IDs, where $n_{in}$ is the number of inputs to that gate.  During each time step of an FSM, a given logic gate produces output as follows.  First, the inputs to the logic gate are converted in order to a binary number $r$ that is used to index the corresponding $r$'th integer in the truth table segment of that gene.  The $n_{out}$ low-order bits of that integer are then isolated, where $n_out$ is the number of outputs from that gate, and these binary values are assigned in order to each output state variable.

\CalloutTable{t:logic-gate} shows an example evolved logic gate, where $X$ and $Y$ are binary inputs, $Z$ is a binary output.  This truth table implements the logic function $Z=X \lor \lnot Y$, illustrating the ability of evolution to use non-standard logic functions.  In all cases, input to-- and output from these gates are binary.  

\begin{table}
    \centering
    \caption{Example truth table for an evolved logic gate.}
    \begin{tabular}{ll|l}
    \hline\hline
        {\bf X} & {\bf Y} & {\bf Z} \\\hline
        0 & 0 & 1\\
        0 & 1 & 0\\
        1 & 0 & 1\\
        1 & 1 & 1\\
        \hline
    \end{tabular}
    \label{t:logic-gate}
\end{table}

\begin{figure}
\centering
\includegraphics[width=2in]{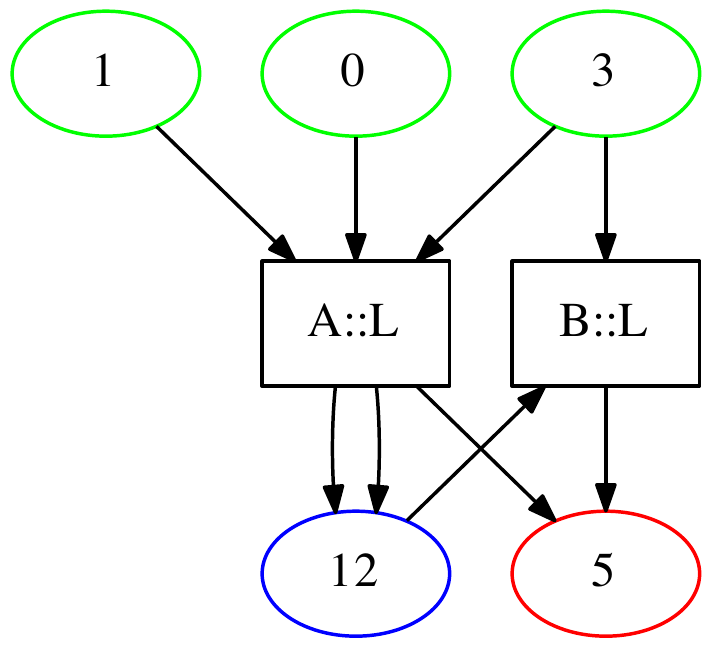}
\caption{Example evolved finite state machine.  This FSM uses two logic gates (black squares, middle) to accept state input (green ovals, top) from three cells in the CA, and produce the state of the cell at the next time step (red oval, bottom right).  One hidden state (blue oval, bottom left) is also used to implement memory.  Gates are labeled with a unique code, while state variables are labeled with their index.}
\label{f:fsm-example2}
\end{figure}

\CalloutFigure{f:fsm-example2} is a high-level depiction of an evolved FSM.  Here we see three inputs (green ovals; top) that are used by two evolved logic gates (black squares; middle).  These inputs are the states of three neighboring cells.  One of these gates (A::L, middle left) produces output that is fed into a hidden state variable (blue oval, bottom left).  This hidden state variable is in turn used by the other logic gate (B::L, middle right), and both logic gates produce output for the cell's next state (red oval, bottom right).  When multiple gates produce output into the same state variable, the values are logically ORed together.

Each genome in the initial population is a random sequence seeded with 16 randomly generated genes.  The entire genome is subject to mutation, including point-mutations (i.e., replacing a codon with a random integer), insertions (i.e., inserting a duplicate of a random sequence from the genome), and deletions (i.e., deleting a random sequence from the genome). Mutations to the underlying genome may alter the truth table, fan-in, and fan-out of the gate, or ``rewire''  the gate to connect to a different state variables.

\subsection{Parameters of the evolutionary algorithm}
The specific techniques and parameters that we used to evolve FSMs are summarized in \CalloutTable{t:params}.  Each individual in a population of 500 was restricted to a genome of size $[1000..40,000)$ integers, which allows up to a few thousand logic gates (in practice, the largest FSM we have observed here comprises only 120 logic gates).  During each update of the EA, the lower 10\% of the population was removed via rank-order selection.  Parents were selected for asexual reproduction proportional to their fitness until the population reached its initial size, and all offspring undergo mutation.  On average, 30 replicates of each treatment were allowed {10,000} updates, which is on average {1,000} generations.  Finally, we note that the sizes of the CAs used during evolution are relatively modest; as will be seen, this does not hamper scalability.

\begin{table}
    \centering
    \caption{Summary of evolutionary algorithm parameters.}
    \begin{tabular}{ll}
    \hline\hline
    {\bf Parameter} & {\bf Value}\\\hline
    Population size & 500\\
    Initial genome size & {10,000}\\
    Minimum genome size & {1,000}\\
    Maximum genome size & {40,000}\\
    Reproduction & asexual\\
    Replacement rate & 0.1\\
    Mutation rate (per-site) & 0.01\\
    Mutation rate (indel) & 0.05\\
    Insertion-deletion size & [16..512)\\
    Number of Initial gates & 16\\
    Samples per fitness evaluation & 100\\
    Fitness averaging window size & 10\\
    Number of updates & {10,000}\\
    Number of generations (mean) & 1000\\
    1D CA topology & $N=35, r=2$\\
    2D CA topology & $(M,N)=(7,7), r=2$\\
    3D CA topology & $(M,N,P)=(3,3,5), r=1$\\
    \hline
    \end{tabular}
    \label{t:params}
\end{table}

\section{Experiments and Results}\label{s:results}

\subsection{Density classification in 1-, 2-, and 3D}

\begin{figure}
\centering\includegraphics[width=\HalfPage]{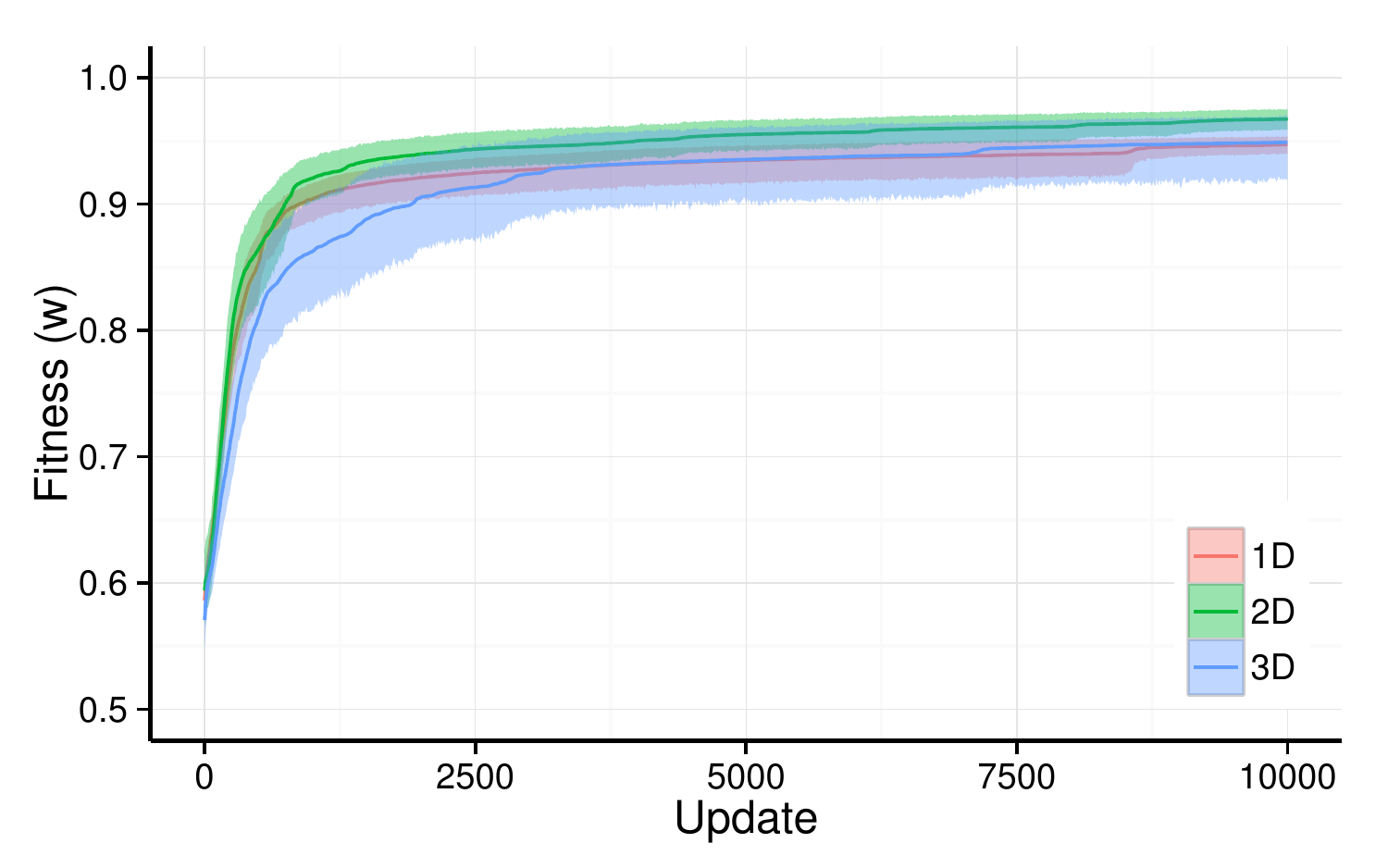}
\caption{Average maximum fitness over time for 1-, 2-, and 3D cellular automata solving the density classification problem.  Shaded regions are 95\% confidence intervals.  The 10,000 updates shown here correspond to, on average, {1,000} generations of evolution.}
\label{p-fitness}
\end{figure}

\CalloutFigure{p-fitness} shows the maximum fitness over time on the density classification task for 1-, 2-, and 3D CA (results are averaged across 30 replicates for each treatment; shaded regions indicate 95\% confidence intervals constructed via {1,000}-fold bootstrapping).  
Fitness improves rapidly over evolutionary time, eventually reaching a level of approximately 0.95 meaning that, on average, the 95\% of the initial configurations are correctly classified (we note that this value is a moving average, as described in \CalloutSection{s:fitness-function}).  This result indicates that our approach to discovering FSM-based update functions for CA using evolvable Markov networks is effective, and can be applied to CA with dramatically different underlying topologies.

\begin{figure}
\centering\includegraphics[width=\HalfPage]{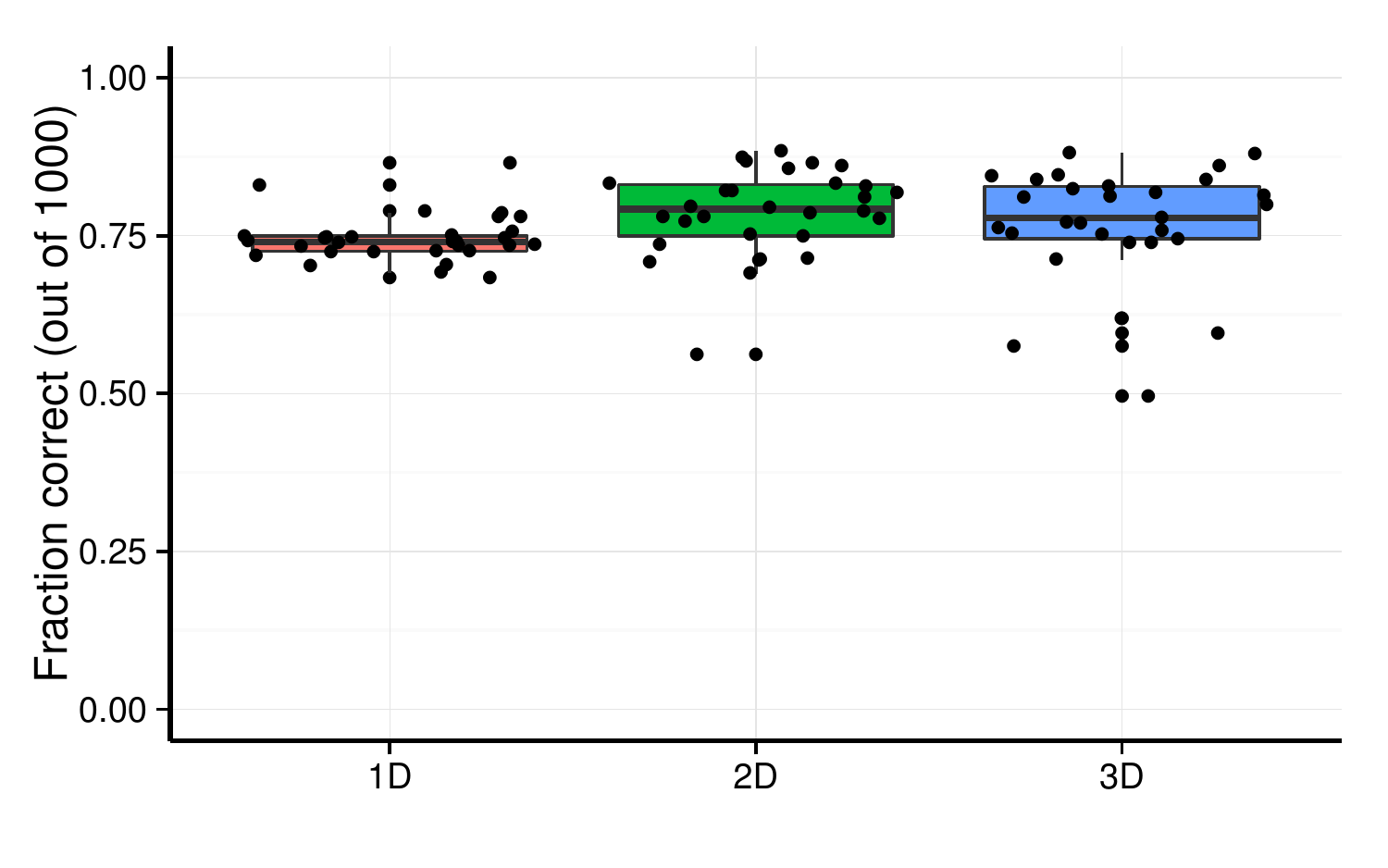}
\caption{Boxplot showing the fraction of correctly classified initial configurations (out of 1000) for the dominant evolved solutions from each treatment (30 replicates per treatment).  Dots are (jittered) values from each replicate.}
\label{f:1000x-fitness}
\end{figure}

To further characterize the performance of CA in classifying initial densities generated by the approach described here, we next analyze the {\em dominants}, which are the most-fit candidate solutions (i.e., FSM specified update functions) at the end of each replicate (30 dominants per treatment).  \CalloutFigure{f:1000x-fitness} shows the fraction of correctly classified initial configurations for each dominant when tested on {1,000} random initial configurations drawn from a binomial distribution ($p=0.5$).  The best dominants from each treatment correctly classified 86.5\%, 88.4\%, and 88.1\% of the initial configurations for the 1-, 2-, and 3D treatments, respectively. This result further demonstrates that evolved update functions perform well even under the challenging 3D topology.

\subsection{Example evolved solutions}

\begin{figure}
\centering\includegraphics[width=.40\textwidth]{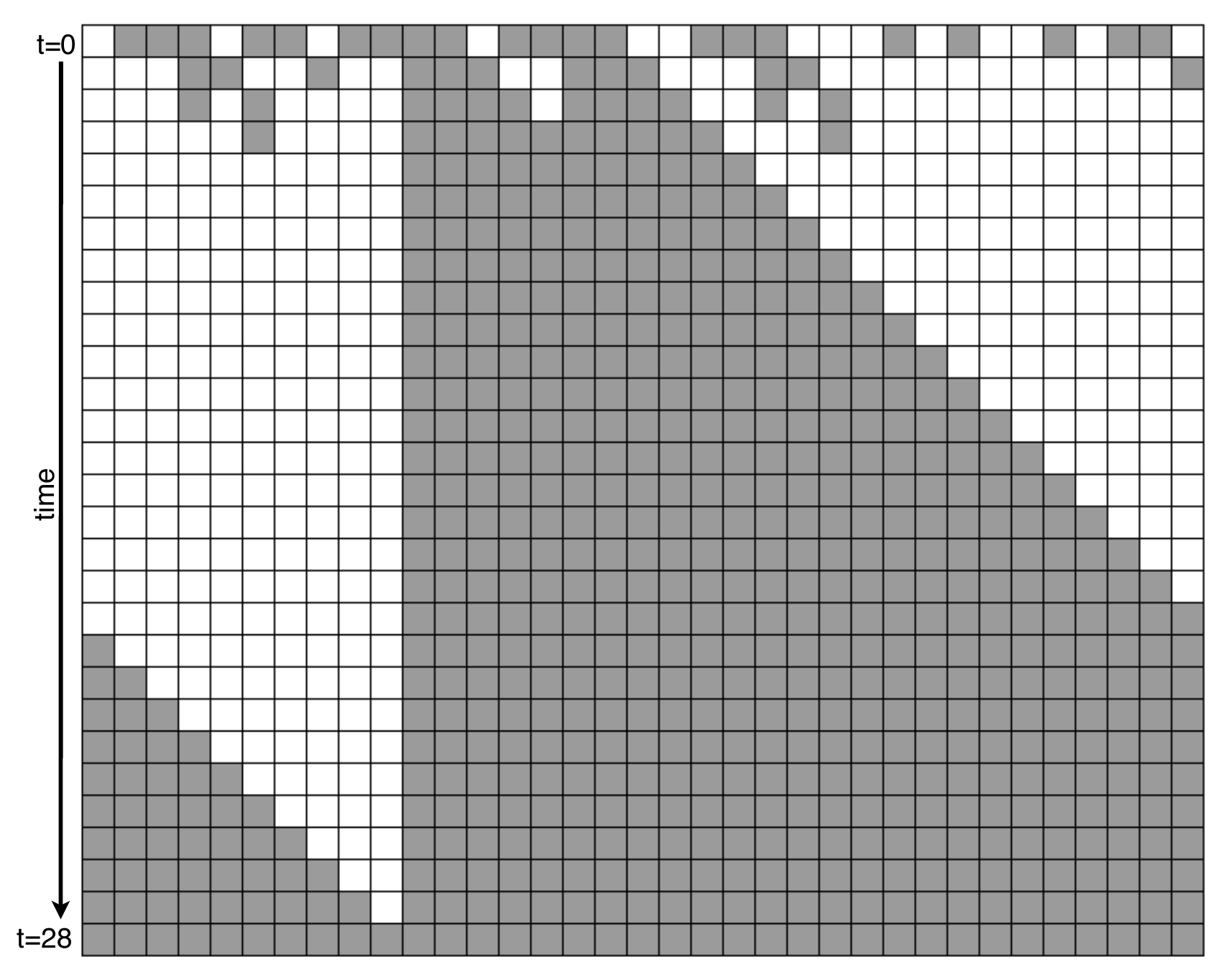}
\caption{Example behavior of a 1D CA.  Each row in this figure represents the configuration of the CA at a single point in time, with time advancing down (top row is $t=0$).  The final configuration (all cells 1), represents the correct classification for this initial configuration.}
\label{f:002-detail}
\end{figure}

\CalloutFigure{f:002-detail} shows an example of the behavior of a dominant (high fitness) rule drawn from one replicate population of the 1D treatment. We show the state of each cell over time during the evaluation of a single initial configuration. 
The CA begins with a majority of black cells. For each time step, all cells execute their FSM to determine their next state. In this case, the cells self-organize from an initially random pattern and transition to an all-black configuration.  This illustration shows how information from different regions of the configuration are used to produce coherent changes over time.

\begin{figure}
\centering\includegraphics[width=.35\textwidth]{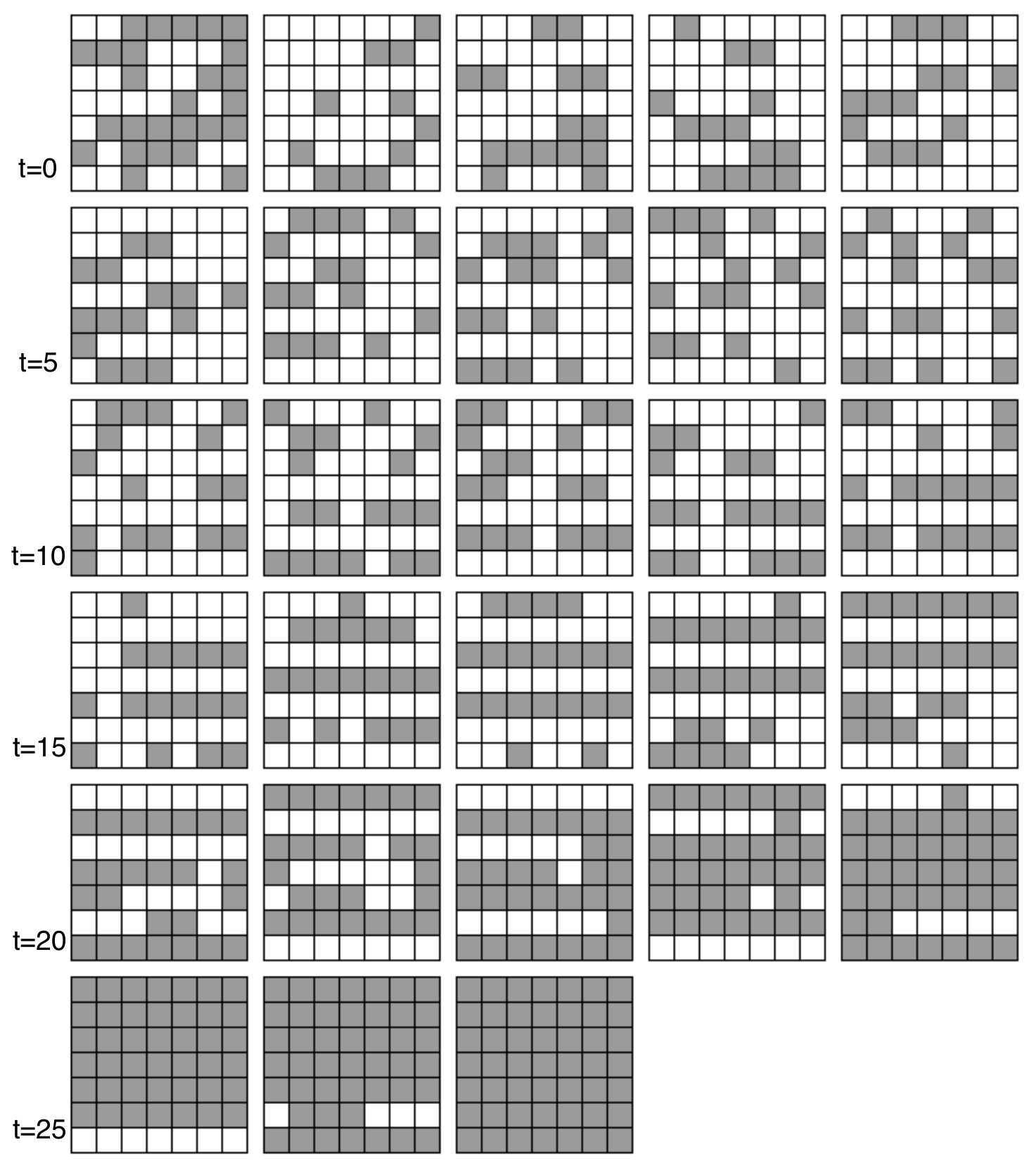}
\caption{Example behavior of a 2D CA.  Each grid in this figure represents the entire state of the CA at a single point in time, with time advancing left-to-right and top-to-bottom (top left is $t=0$).  The final configuration (all cells 1), represents the correct classification for this initial configuration.}
\label{f:003-detail}
\end{figure}

\begin{figure}
\centering\includegraphics[width=.4\textwidth]{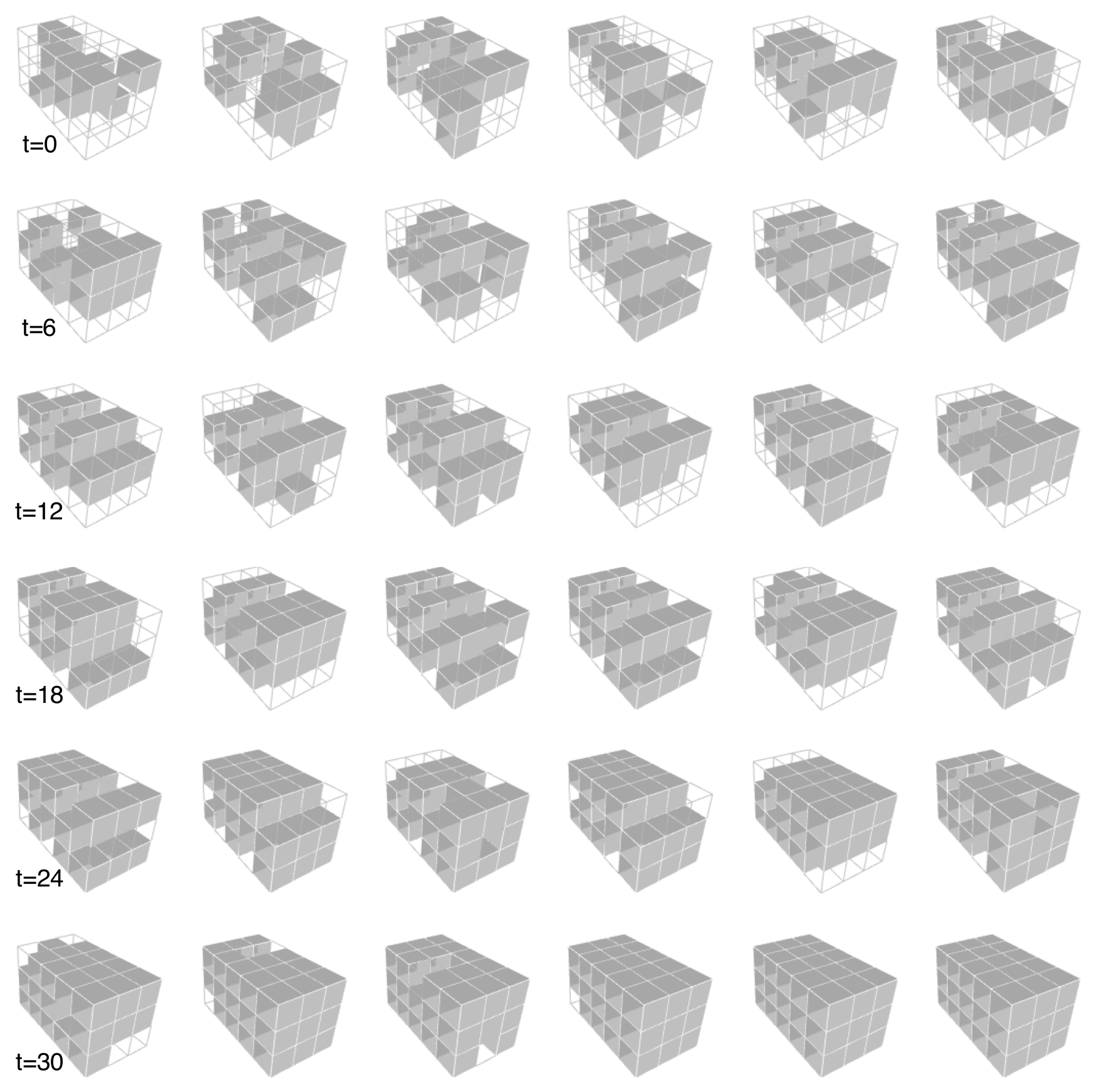}
\caption{Example behavior of a 3D CA.  Each cuboid in this figure represents the entire state of the CA at a single point in time, with time advancing left-to-right and top-to-bottom (top left is $t=0$).  The final configuration (all cells 1), represents the correct classification for this initial configuration.}
\label{f:004-detail}
\end{figure}

Similarly, \CalloutFigure{f:003-detail} depicts the configurations over time during the evaluation of one 2D initial configuration, and \CalloutFigure{f:004-detail} shows configurations for one 3D initial configuration.  As with the 1D case, both figures show the behavior of the dominant individual drawn from the end of a single replicate population and demonstrate the ability of the cells to self-organize towards the correct decision.

\subsection{Rule density}

An important characteristic of CA update functions for density classification task is that the {\em rule density}, which is the average frequency of 1s that are output by an update function,  should be the same as $\rho_c$, the critical density threshold for classification~\cite{Capcarrere:2001kg}.  Thus, we can assess the efficacy of our approach to evolving FSMs to specify update functions by comparing the evolved rule density to the ideal rule density identified by theory.
While this is straightforward to calculate for update functions that are based on binary rule tables, calculating the rule density for evolved FSMs requires that we evaluate the FSM for every possible state, and tally the number of 1s that are output.  In this study, the FSMs for our 1D CA have 22 binary state variables, yielding $2^{22}$ possible states (5 binary state variables for the neighborhood inputs, 1 for the output, and 16 available hidden state variables).  While we are able to calculate the exact rule density for 1D CA, the 2D and 3D CA have $2^42$ and $2^44$ possible states, respectively.  At present, it is computationally infeasible to calculate the exact rule density for 2- and 3D CA, however we were able to estimate rule density by an unbiased sampling of 1 million states.

\begin{figure}
\centering\includegraphics[width=\HalfPage]{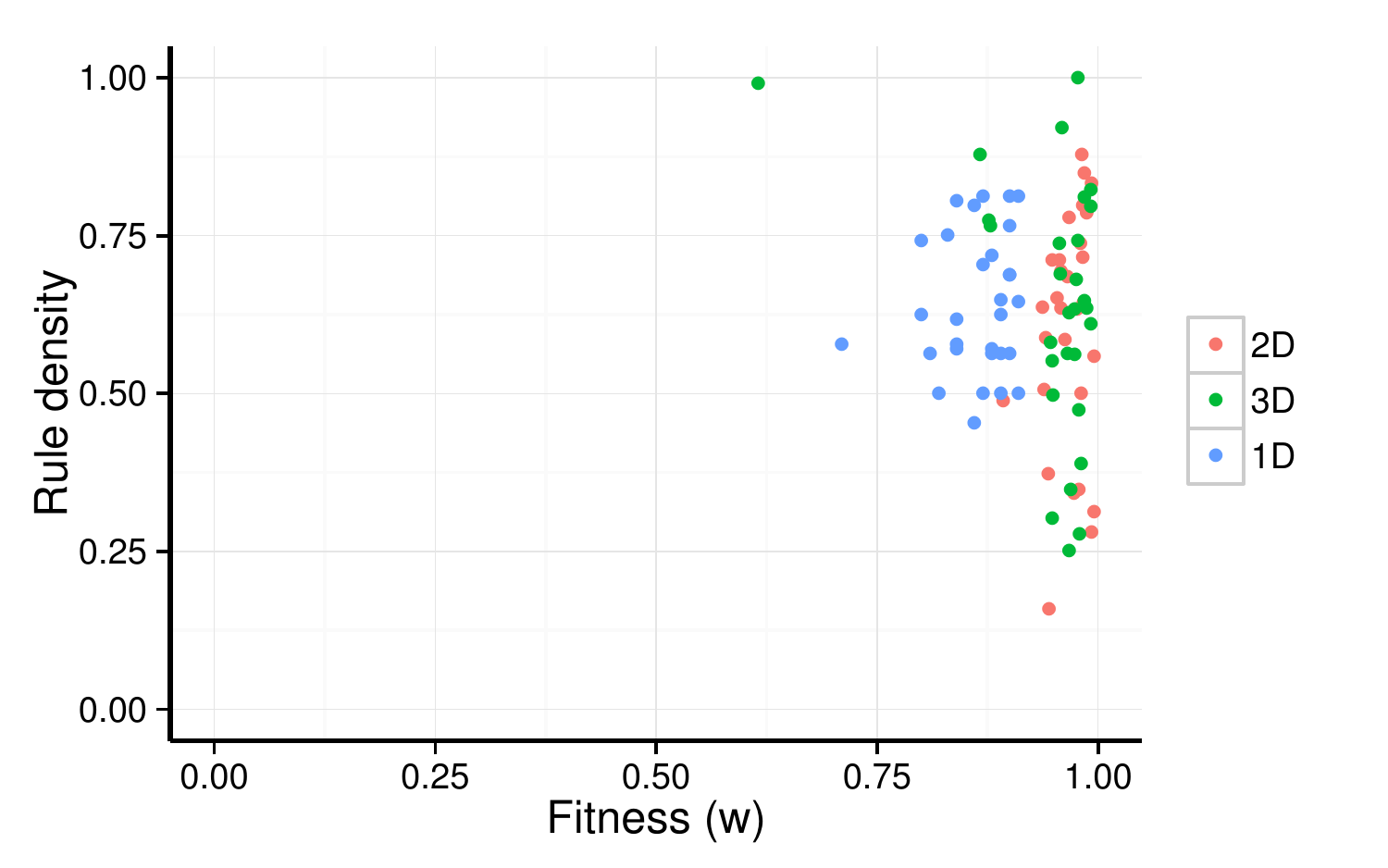}
\caption{Rule density density for the evolved dominant solutions from each treatment.  1D results are exact, while 2- and 3D rule densities are estimated from 1M random samples.}
\label{f:rule-density}
\end{figure}

We show in \CalloutFigure{f:rule-density} the fitness of dominant solutions vs.~rule density for each of 1-, 2-, and 3D CA.  In general, we observe that many of the evolved update functions do indeed have a rule density approaching $1/2$, however others do not exhibit this characteristic.  Furthermore, we find little correlation between rule density and fitness (Spearman rank correlation, $\rho=0.106$).  We conjecture that the use of memory (made possible by the use of hidden states) alters the optimal density of the rule table, a hypothesis that we hope to explore in future work.  Finally, a number of solutions appear to have densities which belie their relatively high fitness.  We note that (1) the states over which we calculate rule density are not necessarily the same states that are {\em actually visited} during evaluation of an initial configuration, and (2) we have observed scenarios where nearly the entire configuration approaches 1, and then in a coordinated fashion turns completely 0 in a single time step.  The density for the rule table corresponding to this behavior would be predominately 1, with perhaps a few 0s.  This intriguing behavior is only possible in the presence of memory, and also warrants further investigation.

\subsection{Self-adaptation}
Self-adaptation (in the present context) is the unsupervised usage of information that elicits a (run-time) behavioral change in an agent. In general, this requires the capacity to sense environmental characteristics, and to modify internal parameters according to the sensed values. Considering that most CA use an update function defined by a fixed binary rule table (and thus do not include memory), traditional CA are not capable of self-adaptation.  However, cells in a CA with memory are in theory capable of using these internal states to change their behavior in response to local stimuli.  

\begin{figure}
\centering\includegraphics[width=\HalfPage]{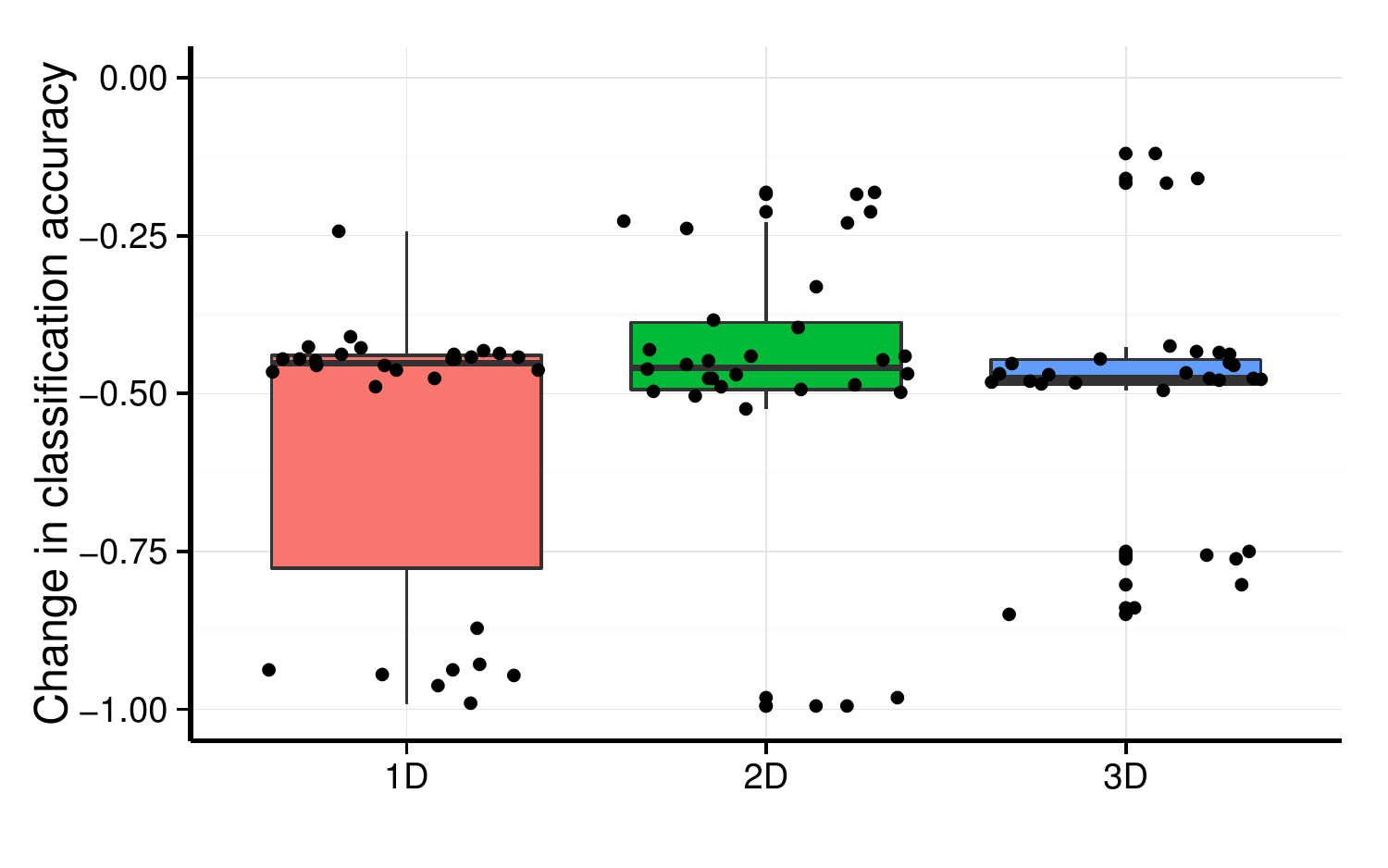}
\caption{Boxplot showing the effect of hidden states (memory) on the fraction of correctly classified initial configurations for 1-, 2-, and 3D cellular automata.}
\label{f:ko-hidden}
\end{figure}

The FSMs evolved here include 16 hidden state variables.  While we do not predispose our evolutionary algorithm to use any of these hidden states, they could in principle be used as a form of memory for an update function.  To determine if memory was an important part of evolved solutions, we evaluated all dominants on {1,000} random (binomial) initial configurations where we held the values of all hidden state variables at 0.  We show in \CalloutFigure{f:ko-hidden} how the fraction of correctly classified $IC$s ($\Delta w$) changes for each of the 1-, 2-, and 3D CA.  As seen there, the average decline in classification accuracy is approximately 0.5 for all treatments.  This result strongly indicates that not only are the hidden state variables used, they positively contribute to the performance of evolved FSMs. Because the internal states must depend on the particular $IC$ the automaton was presented with, we can conclude that the evolved FSMs are self-adaptive. 

\subsection{Self-organization}
Systems where individuals react to each other locally without the availability of global information (as in CA) are examples of {\em self-organization}.  There are various definitions of self-organization, for example:  ``Self-organization is a process in which pattern at the global level of a system emerges solely from numerous interactions among the lower-level components of the system,'' provided by Camazine~\cite{Camazine:2003tb}.  While self-organization can be challenging to quantify, we have previously defined {\em operational self-organization}, $S_op(x)$, as:
\begin{equation}
S_{op}(x) =\frac{f(x)-f_{nc}(x)}{f(x)+f_{nc}(x)}
\label{e:self-org}
\end{equation}
where $f(x)$ is a performance metric of system $x$ with communication among agents enabled, and $f_{nc}(x)$ is the same performance metric of $x$ with communication disabled~\cite{Knoester:2011gp}.  $S_{op}(x)$ has the interesting property that values greater than zero indicate the presence of self-organization (communication is beneficial), while values less than or equal to zero indicate the lack of self-organization (communication is harmful or neutral).  Here, we can study the degree of self-organization by disabling communication among cells, which is achieved by holding inputs from neighbors (non-self) to 0. 

\begin{figure}
\centering\includegraphics[width=\HalfPage]{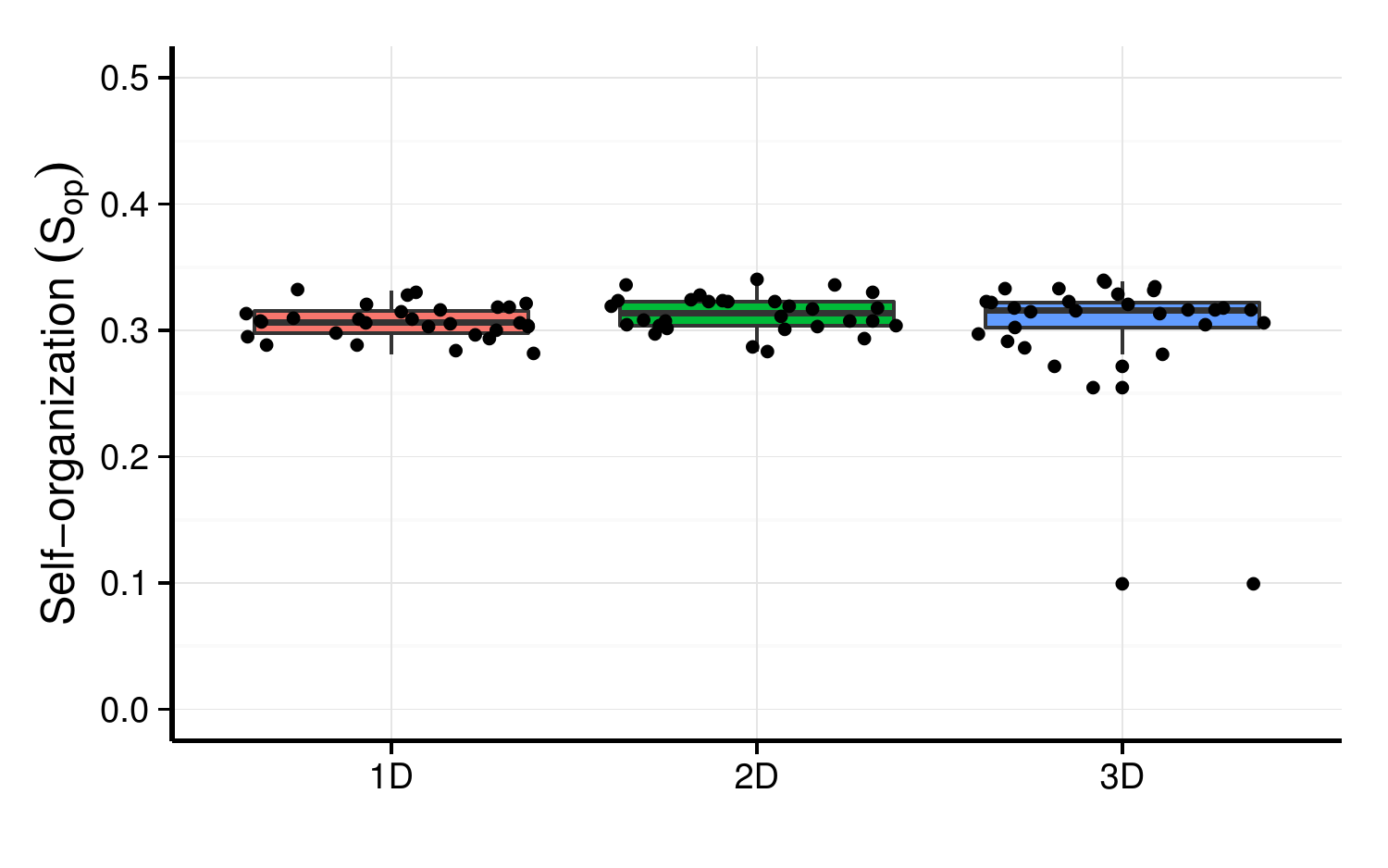}
\caption{Boxplot showing operational self-organization for the dominant evolved FSMs from each treatment.  Solutions strongly exhibit self-organization.}
\label{f:ko-input}
\end{figure}

\CalloutFigure{f:ko-input} is a boxplot showing $S_{op}(x)$ for the dominant solutions from each treatment.  As shown there, $S_{op}(x)$ is positive, which indicates that self-organization is always beneficial.  Interestingly, $S_{op}(x)$ is not 1.0; this indicates that in many cases, the evolved FSM correctly classifies an initial configuration without communication among cells.  This is an artifact of both our mechanism for disabling communication (holding inputs at 0) and of evolved FSMs having a steady-state behavior that happened to match the correct classification for a given random initial configuration.

\subsection{Scalability}
\begin{figure}
\centering\includegraphics[width=\HalfPage]{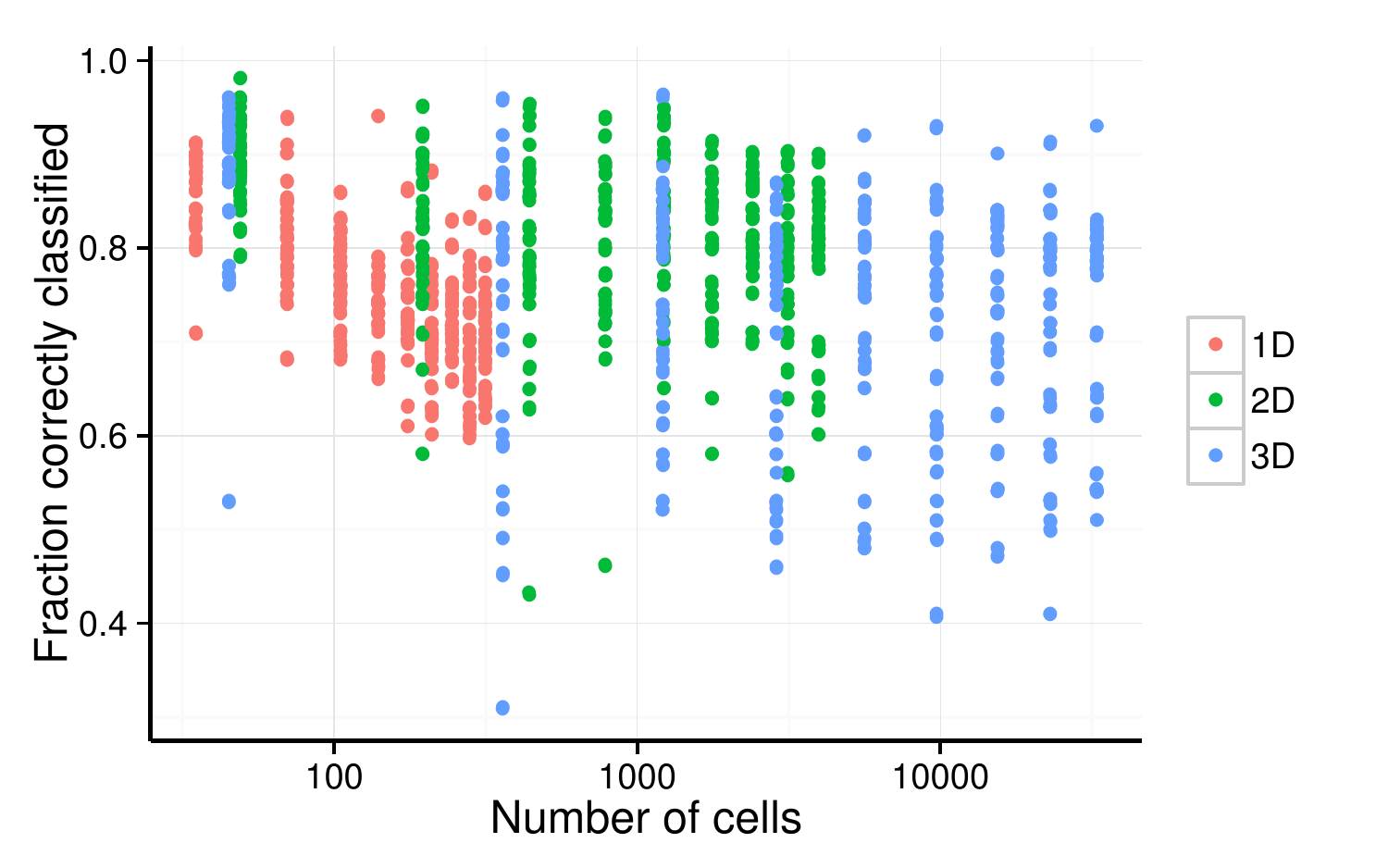}
\caption{Each dominant FSM is highly scalable, which we show by testing them in larger CA.  We scale each dimension of the 1-, 2-, and 3D CA by a factor $s$ that ranges from $[0..9]$, and plot the fraction of correctly classified configurations vs.~the total number of cells.  Note that the $x$-axis has a log-scale.}
\label{f:scalability}
\end{figure}

An important and desirable characteristic of SASO systems is that they exhibit a high degree of scalability~\cite{Knoester:2011gp}.  To explore the scalability of the evolved FSMs, we multiplied each dimension of the 1-, 2-, and 3D CA by a scaling factor $s=[1..9]$ and re-evaluated the dominant individuals from each treatment on {1,000} new random initial configurations.  We can see in \CalloutFigure{f:scalability} the resulting fraction of correctly classified random initial configurations (out of {1,000}; we note that the $x$-axis is log-scale).  We note that the evolved solutions are highly scalable, in some cases scaling multiple orders of magnitude above the input size that they had evolved with, with little decline in performance.  Compared to other studies, these results are quite competitive.  For example, in~\cite{Mitchell:1996ut}, Mitchell~{\etal} describe an evolved solution for a 1D $N=149$ CA that achieves a classification accuracy of 76.9\%, while Andre~{\etal} achieve an accuracy of 82.33\%, again in 1D~\cite{Andre:1996wa}.  In contrast, depending on the size and topology of the CA, the solutions we discover here surpass 95\%.

\begin{figure}
\centering\includegraphics[width=\HalfPage]{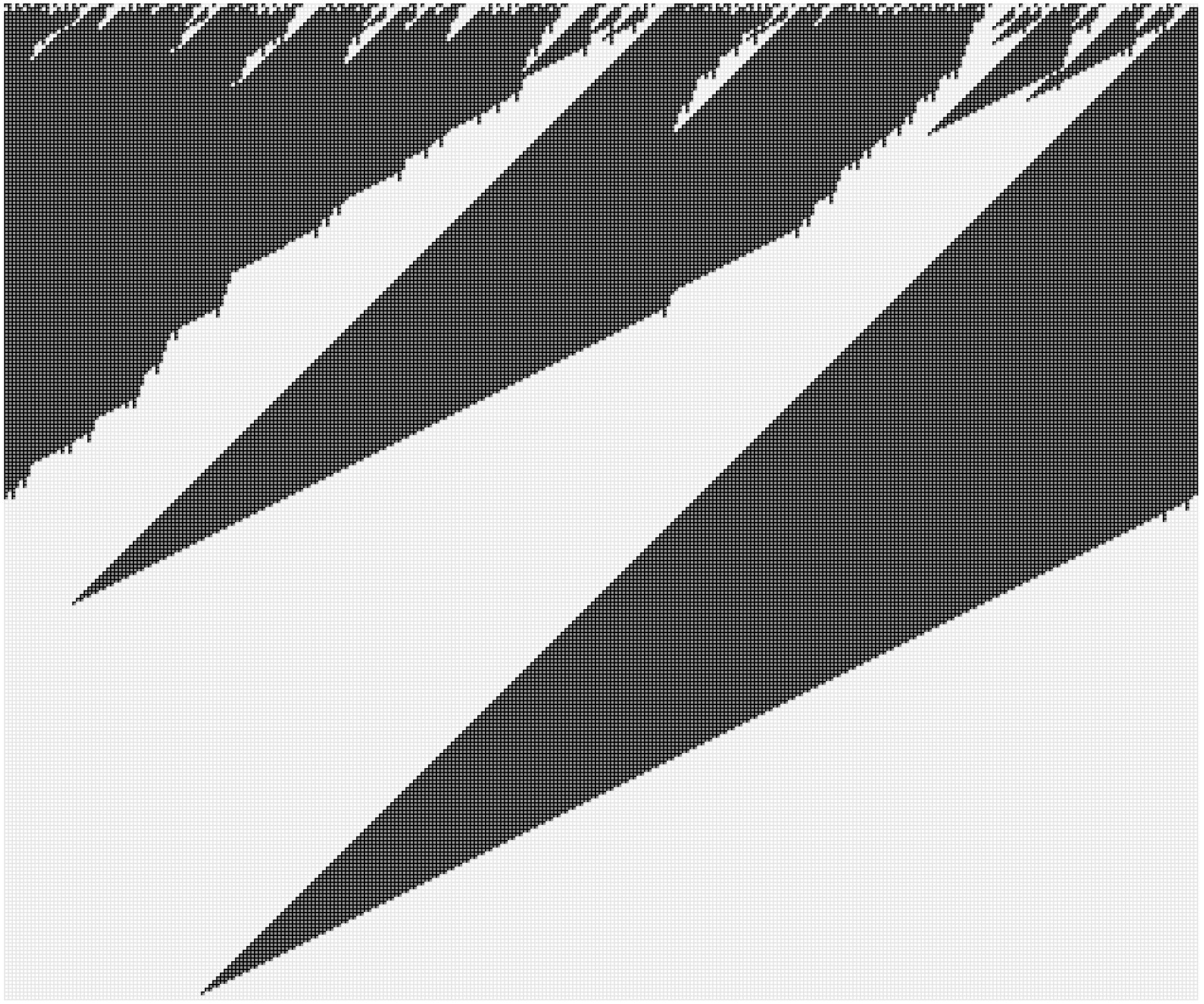}
\caption{Example behavior of an evolved FSM on a 1D CA evaluated on a configuration of 315 cells (9 times larger than during evolution).}
\label{f:example-1d-large}
\end{figure}

\begin{figure}
\centering\includegraphics[width=\HalfPage]{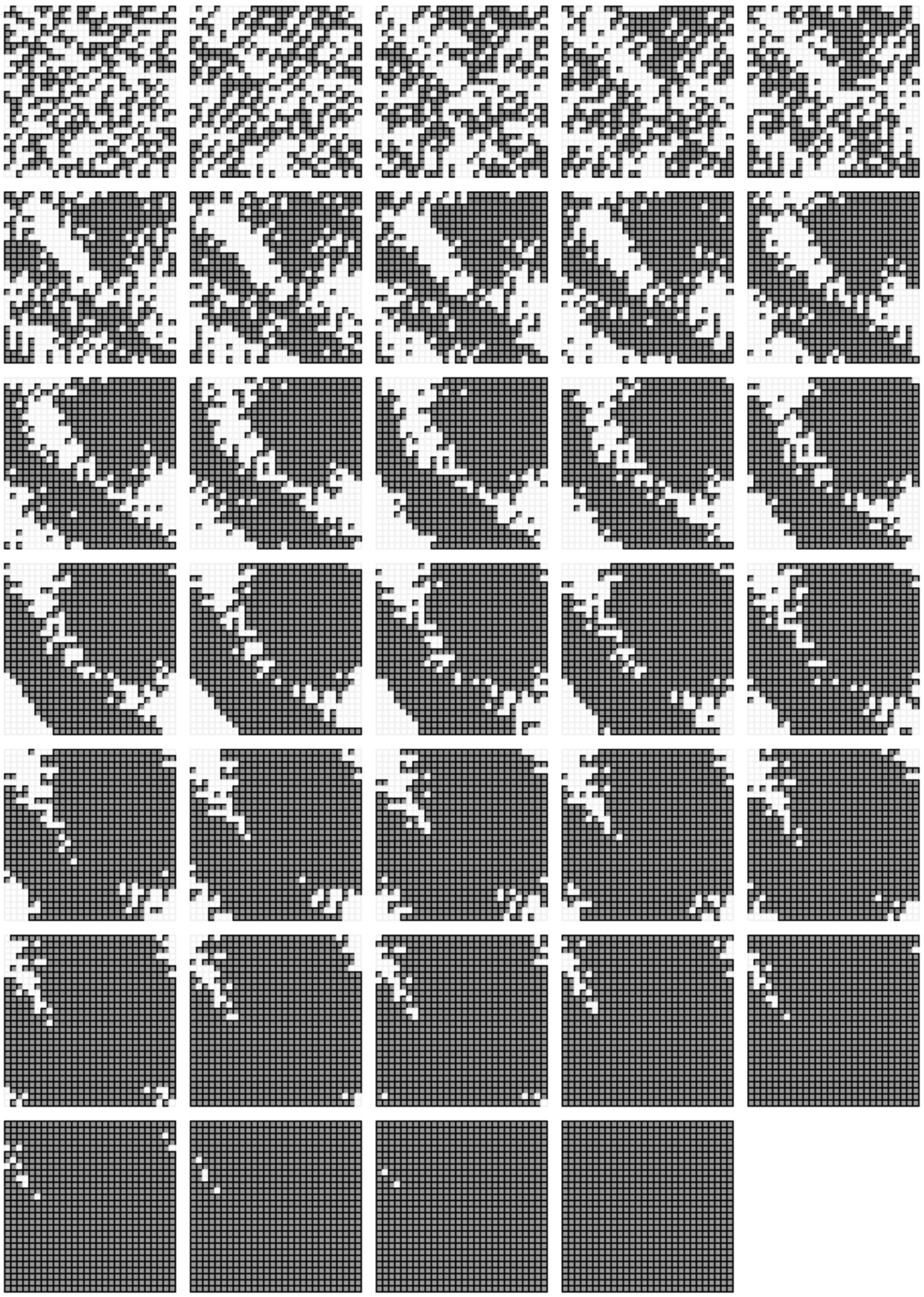}
\caption{Example behavior of an evolved FSM on a 2D CA evaluated on a configuration of 784 cells (16 times larger than during evolution).}
\label{f:example-2d-large}
\end{figure}

Finally, \CalloutFigure{f:example-1d-large} and \CalloutFigure{f:example-2d-large} document the time series of configuration changes for example 1D and 2D CA that have been scaled by 9 and 4, respectively.  As seen here, the patterns produced by the evolved FSMs are quite intriguing.  While difficult to see within one figure, the animations corresponding to these behaviors indicate that {\em particles}, that is, cohesive moving blocks of black cells, readily form in these configurations~\cite{Mitchell:1996ut}.

\subsection{Example evolved FSMs}
\begin{figure}
\centering
\subfloat[1D]{\includegraphics[width=.4\textwidth]{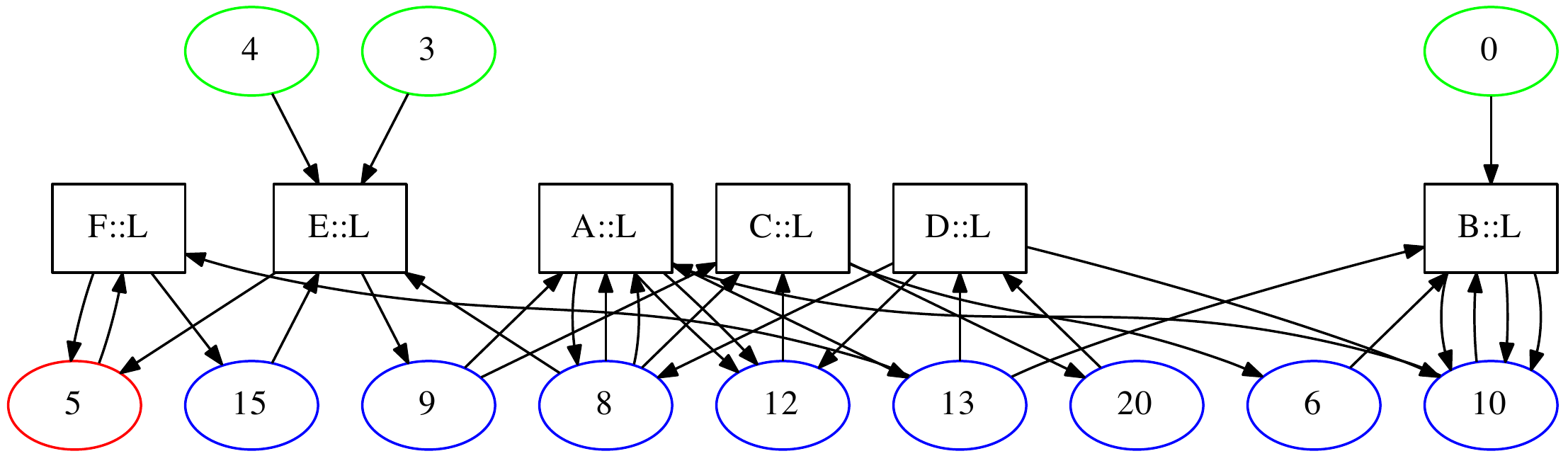}\label{f:big-1}}
\hfil
\subfloat[2D]{\includegraphics[width=.4\textwidth]{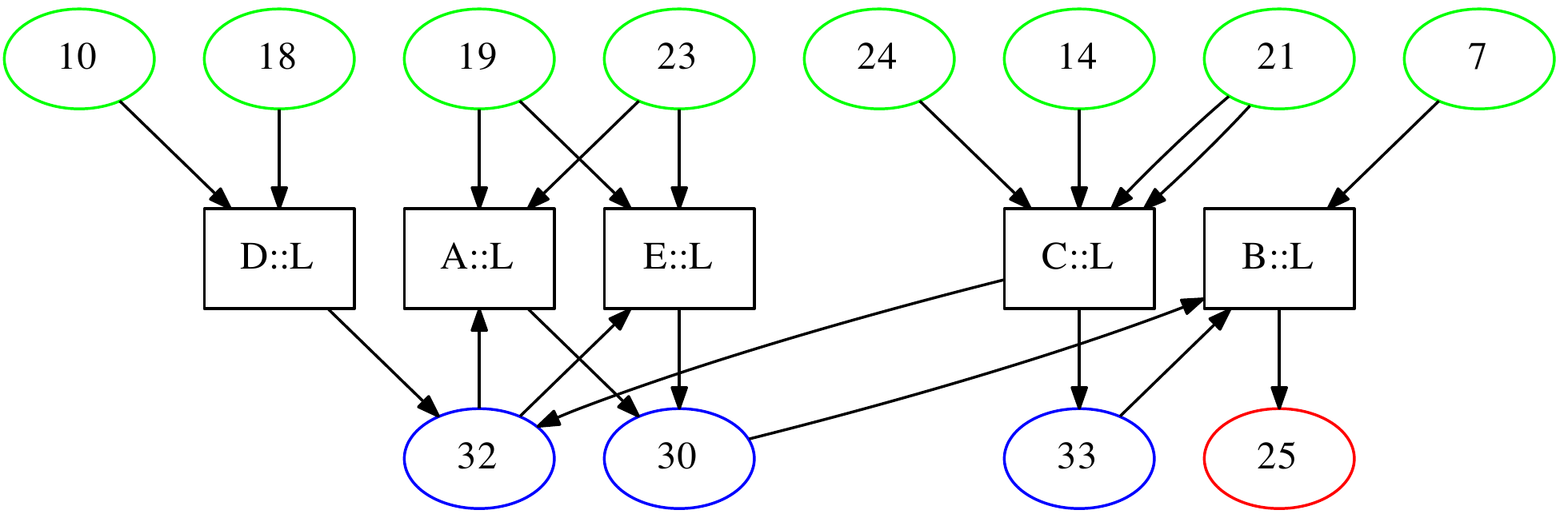}\label{f:big-2}}
\hfil
\subfloat[3D]{\includegraphics[width=.4\textwidth]{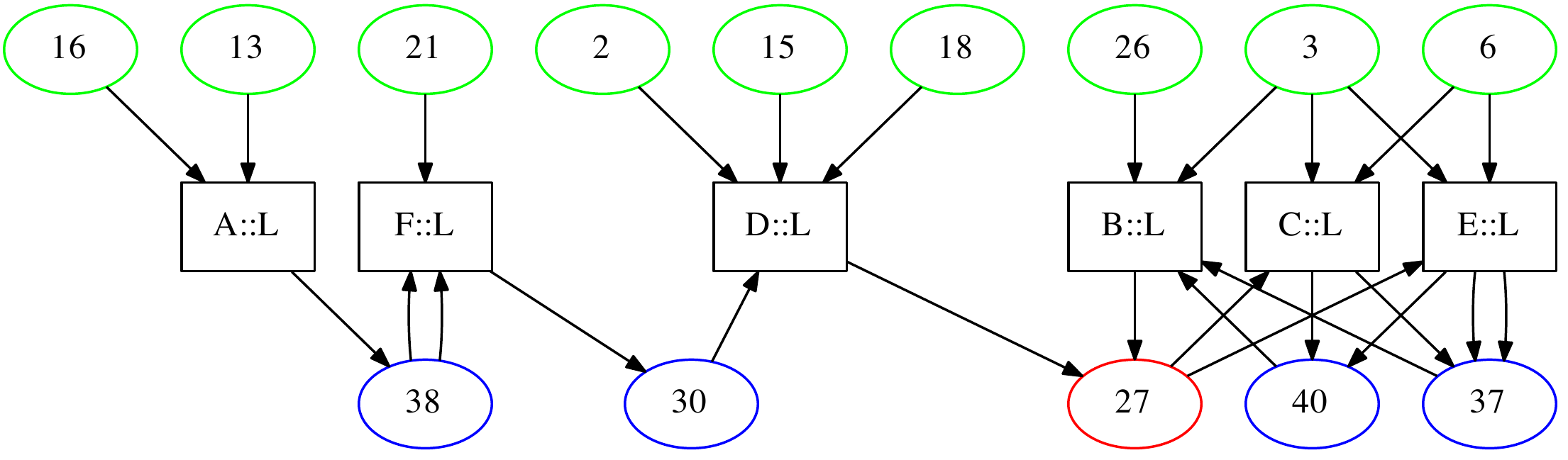}\label{f:big-3}}
\caption[]{Evolved FSM connection diagrams for 1-, 2-, and 3D CA.  Shown here are the logic gates (black squares), input states (green ovals), and output states (red ovals), and hidden states (blue ovals).  Gates are labeled with a unique character for tracking purposes only, while state variables are labeled with their index.}
\label{f:big-fsm-examples}
\vspace{1em}
\end{figure}

We show in \CalloutFigure{f:big-fsm-examples} the FSM connection diagrams for each of a 1-, 2-, and 3D CA that achieved fitness scores slightly over 0.99 (the optimal fitness is 1.0).  As seen here, these FSMs evolved to use between 3 and 8 hidden and input states.  In all cases, the number of inputs used is below that available in each cell's neighborhood, while the memory afforded each cell by the inclusion of hidden states improved their classification accuracy.  These characteristics of the evolved FSMs highlight the ability of EAs to discover non-intuitive solutions to a challenging problem.
 
\section{Conclusion}\label{s:conclusion}
We have described an approach in which evolutionary algorithms can be used to discover finite state machines, and used this technique to discover FSMs that perform the density classification task in cellular automata.  Moreover, we have shown that CA that use these evolved FSMs are self-organizing, self-adaptive, and highly scalable.  In future work, 
we plan to extend this approach to include other aspects of {\selfstar} systems, and explore the discovery of FSMs that are fault-tolerant and self-healing. 

\bibliographystyle{IEEEtran}
\bibliography{references}

\begin{thebibliography}{10}
\providecommand{\url}[1]{#1}
\csname url@samestyle\endcsname
\providecommand{\newblock}{\relax}
\providecommand{\bibinfo}[2]{#2}
\providecommand{\BIBentrySTDinterwordspacing}{\spaceskip=0pt\relax}
\providecommand{\BIBentryALTinterwordstretchfactor}{4}
\providecommand{\BIBentryALTinterwordspacing}{\spaceskip=\fontdimen2\font plus
\BIBentryALTinterwordstretchfactor\fontdimen3\font minus
  \fontdimen4\font\relax}
\providecommand{\BIBforeignlanguage}[2]{{%
\expandafter\ifx\csname l@#1\endcsname\relax
\typeout{** WARNING: IEEEtran.bst: No hyphenation pattern has been}%
\typeout{** loaded for the language `#1'. Using the pattern for}%
\typeout{** the default language instead.}%
\else
\language=\csname l@#1\endcsname
\fi
#2}}
\providecommand{\BIBdecl}{\relax}
\BIBdecl

\bibitem{Wolf:2009uh}
W.~Wolf, ``{Cyber-physical Systems},'' \emph{IEEE Computer}, vol.~42, no.~3,
  2009.

\bibitem{Kephart:2003uq}
J.~O. Kephart and D.~M. Chess, ``{The Vision of Autonomic Computing},''
  \emph{IEEE Computer}, vol.~36, pp. 41--50, 2003.

\bibitem{Babaoglu:2005wb}
O.~Babaoglu, M.~Jelasity, A.~Montresor, C.~Fetzer, S.~Leonardi, A.~v. Moorsel,
  and M.~v. Steen, \emph{{Self-star Properties in Complex Information Systems:
  Conceptual and Practical Foundations (Lecture Notes in Computer
  Science)}}.\hskip 1em plus 0.5em minus 0.4em\relax Secaucus, NJ, USA:
  Springer-Verlag New York, Inc., 2005.

\bibitem{Holldobler:2008ws}
B.~H{\"o}lldobler and E.~O. Wilson, \emph{{The Superorganism: The Beauty,
  Elegance, and Strangeness of Insect Societies}}.\hskip 1em plus 0.5em minus
  0.4em\relax W. W. Norton {\&} Company, 2008.

\bibitem{Bradshaw:1965uh}
A.~D. Bradshaw, ``{Evolutionary Significance of Phenotypic Plasticity in
  Plants},'' \emph{Advances in genetics}, 1965.

\bibitem{WestEberhard:1989wq}
M.~J. West-Eberhard, ``{Phenotypic plasticity and the origins of diversity},''
  \emph{Annual review of ecology and systematics}, 1989.

\bibitem{Agrawal:2001ji}
A.~A. Agrawal, ``{Phenotypic Plasticity in the Interactions and Evolution of
  Species},'' \emph{Science}, vol. 294, no. 5541, pp. 321--326, Oct. 2001.

\bibitem{Kramer:2010fg}
O.~Kramer, ``{Evolutionary self-adaptation: a survey of operators and strategy
  parameters},'' \emph{Evolutionary Intelligence}, vol.~3, no.~2, pp. 51--65,
  Feb. 2010.

\bibitem{Gardner:1970to}
M.~Gardner, ``{Mathematical games: The fantastic combinations of John Conway's
  new solitaire game ``life''},'' \emph{Scientific American}, 1970.

\bibitem{Neumann:1966wc}
J.~V. Neumann and A.~W. Burks, \emph{{Theory of Self-Reproducing
  Automata}}.\hskip 1em plus 0.5em minus 0.4em\relax University of Illinois
  Press, 1966.

\bibitem{Holland:1970fy}
J.~H. Holland, ``{Logical Theory of Adaptive Systems},'' in \emph{Essays on
  Cellular Automata}, A.~W. Burks, Ed.\hskip 1em plus 0.5em minus 0.4em\relax
  University of Illinois Press, 1970.

\bibitem{Wolfram:1994tc}
S.~Wolfram, \emph{{Cellular Automata and Complexity: Collected Papers}}.\hskip
  1em plus 0.5em minus 0.4em\relax Addison-Wesley, 1994.

\bibitem{Cook:2004tt}
M.~Cook, ``{Universality in elementary cellular automata},'' \emph{Complex
  Systems}, 2004.

\bibitem{Land:1995ha}
M.~Land and R.~Belew, ``{No Perfect Two-State Cellular Automata for Density
  Classification Exists},'' \emph{Physical review letters}, vol.~74, no.~25,
  pp. 5148--5150, Jun. 1995.

\bibitem{Darabos:2013if}
C.~Darabos, C.~Mackenzie, M.~Tomassini, M.~Giacobini, and J.~Moore, ``{Cellular
  Automata Coevolution of Update Functions and Topologies: A Tradeoff between
  Accuracy and Speed},'' in \emph{European Conference on Artificial Life
  2013}.\hskip 1em plus 0.5em minus 0.4em\relax MIT Press, Sep. 2013, pp.
  340--347.

\bibitem{Mitchell:1996ut}
M.~Mitchell, J.~P. Crutchfield, and R.~Das, ``{Evolving cellular automata with
  genetic algorithms: A review of recent work},'' in \emph{Proceedings of the
  First International {\ldots}}, 1996.

\bibitem{Andre:1996wa}
D.~Andre, F.~H. Bennett, III, and J.~R. Koza, \emph{{Discovery by genetic
  programming of a cellular automata rule that is better than any known rule
  for the majority classification problem}}.\hskip 1em plus 0.5em minus
  0.4em\relax MIT Press, Jul. 1996.

\bibitem{Breukelaar:2005jz}
R.~Breukelaar and T.~B{\"a}ck, ``{Using a genetic algorithm to evolve behavior
  in multi dimensional cellular automata},'' in \emph{the 2005
  conference}.\hskip 1em plus 0.5em minus 0.4em\relax New York, New York, USA:
  ACM Press, 2005, p. 107.

\bibitem{Chavoya:2006uz}
A.~Chavoya and Y.~Duthen, ``{Using a genetic algorithm to evolve cellular
  automata for 2D/3D computational development},'' \emph{{\ldots} of the 8th
  annual conference on Genetic and {\ldots}}, 2006.

\bibitem{Rosin:2006fd}
P.~L. Rosin, ``{Training cellular automata for image processing},'' \emph{IEEE
  Transactions on Image Processing}, vol.~15, no.~7, pp. 2076--2087, 2006.

\bibitem{Reynolds:1987iz}
C.~W. Reynolds, ``{Flocks, herds and schools: A distributed behavioral
  model},'' \emph{ACM SIGGRAPH Computer Graphics}, vol.~21, no.~4, pp. 25--34,
  Aug. 1987.

\bibitem{Langton:1986ja}
C.~G. Langton, ``{Studying artificial life with cellular automata},''
  \emph{Physica D: Nonlinear Phenomena}, vol.~22, no. 1-3, pp. 120--149, Oct.
  1986.

\bibitem{Bandini:2001ch}
S.~Bandini, G.~Mauri, and R.~Serra, ``{Cellular automata: From a theoretical
  parallel computational model to its application to complex systems},''
  \emph{Parallel Computing}, vol.~27, no.~5, pp. 539--553, Apr. 2001.

\bibitem{Sayama:2004bf}
H.~Sayama, ``{Self-Protection and Diversity in Self-Replicating Cellular
  Automata},'' \emph{Artificial Life}, vol.~10, no.~1, pp. 83--98, Jan. 2004.

\bibitem{Maignan:2011cd}
L.~Maignan and F.~Gruau, ``{Convex Hulls on Cellular Spaces: Spatial Computing
  on Cellular Automata},'' in \emph{2011 Fifth IEEE Conference on Self-Adaptive
  and Self-Organizing Systems Workshops (SASOW)}.\hskip 1em plus 0.5em minus
  0.4em\relax IEEE, 2011, pp. 67--72.

\bibitem{AlonsoSanz:2012ub}
R.~Alonso-Sanz, ``{Cellular Automata with Memory},'' \emph{Computational
  Complexity: Theory}, 2012.

\bibitem{Chua:2002vl}
L.~O. Chua and T.~Roska, \emph{{Cellular Neural Networks and Visual Computing:
  Foundations and Applications}}.\hskip 1em plus 0.5em minus 0.4em\relax
  Cambridge University Press, 2002.

\bibitem{Harrer:1992it}
H.~Harrer and J.~A. Nossek, ``{Discrete-time cellular neural networks},''
  \emph{International Journal of Circuit Theory and Applications}, vol.~20,
  no.~5, pp. 453--467, Sep. 1992.

\bibitem{Eiben:2003tf}
A.~E. Eiben and J.~E. Smith, \emph{{Introduction to Evolutionary
  Computing}}.\hskip 1em plus 0.5em minus 0.4em\relax Springer, 2003.

\bibitem{Fogel:1994eg}
D.~B. Fogel, ``{An introduction to simulated evolutionary optimization},''
  \emph{IEEE Transactions on Neural Networks}, vol.~5, no.~1, pp. 3--14, 1994.

\bibitem{Knoester:2008wk}
D.~B. Knoester, P.~K. McKinley, and C.~Ofria, ``{Cooperative Network
  Construction Using Digital Germlines},'' in \emph{Proceedings of the Genetic
  and Evolutionary Computation Conference (GECCO)}, Jul. 2008, pp. 217--224.
\newpage
\bibitem{Knoester:2009vl}
D.~B. Knoester and P.~K. McKinley, ``{Evolution of Probabilistic Consensus in
  Digital Organisms},'' in \emph{Proceedings of the IEEE International
  Conference on Self-Adaptive and Self-Organizing Systems (SASO)}, Sep. 2009.

\bibitem{Knoester:2011gp}
------, ``{Neuroevolution of Controllers for Self-Organizing Mobile Ad Hoc
  Networks},'' in \emph{Proceedings of the IEEE International Conference on
  Self-Adaptive and Self-Organizing Systems (SASO)}.\hskip 1em plus 0.5em minus
  0.4em\relax IEEE, Oct. 2011.

\bibitem{Beckmann:2007tt}
B.~E. Beckmann, P.~K. McKinley, D.~B. Knoester, and C.~Ofria, ``{Evolution of
  Cooperative Information Gathering in Self-Replicating Digital Organisms},''
  in \emph{Proceedings of the International Conference on Self-Adaptive and
  Self-Organizing Systems (SASO)}, Jul. 2007.

\bibitem{Goldsby:2014df}
H.~J. Goldsby, D.~B. Knoester, C.~Ofria, and B.~Kerr, ``{The Evolutionary
  Origin of Somatic Cells under the Dirty Work Hypothesis},'' \emph{PLOS
  Biology}, vol.~12, no.~5, p. e1001858, May 2014.

\bibitem{Stone:2009ki}
C.~Stone and L.~Bull, ``{Evolution of cellular automata with memory: The
  Density Classification Task},'' \emph{Biosystems}, vol.~97, no.~2, pp.
  108--116, Aug. 2009.

\bibitem{Fogel:1966wb}
L.~J. Fogel, A.~J. Owens, and M.~J. Walsh, \emph{{Artificial Intelligence
  through Simulated Evolution}}.\hskip 1em plus 0.5em minus 0.4em\relax John
  Wiley {\&} Sons, Inc., 1966.

\bibitem{Fogel:2006uy}
D.~B. Fogel, \emph{{Evolutionary Computation: Toward a New Philosophy of
  Machine Intelligence}}.\hskip 1em plus 0.5em minus 0.4em\relax John Wiley
  {\&} Sons, Inc., 2006.

\bibitem{Fogel:2008hy}
G.~B. Fogel, ``{Computational intelligence approaches for pattern discovery in
  biological systems},'' \emph{Briefings in Bioinformatics}, vol.~9, no.~4, pp.
  307--316, Mar. 2008.

\bibitem{Fogel:1999vs}
L.~J. Fogel, \emph{{Intelligence through simulated evolution: forty years of
  evolutionary programming}}.\hskip 1em plus 0.5em minus 0.4em\relax John Wiley
  {\&} Sons, Inc., 1999.

\bibitem{Fogel:1991va}
D.~B. Fogel, \emph{{System Identification through Simulated Evolution: A
  Machine Learning Approach to Modeling}}.\hskip 1em plus 0.5em minus
  0.4em\relax Ginn Press, Jul. 1991.

\bibitem{Edlund:2011kt}
J.~A. Edlund, N.~Chaumont, A.~Hintze, C.~Koch, G.~Tononi, and C.~Adami,
  ``{Integrated Information Increases with Fitness in the Evolution of
  Animats},'' \emph{PLoS Computational Biology}, vol.~7, no.~10, p. e1002236,
  Oct. 2011.

\bibitem{Olson:2013kx}
R.~S. Olson, A.~Hintze, F.~C. Dyer, D.~B. Knoester, and C.~Adami, ``{Predator
  confusion is sufficient to evolve swarming behaviour},'' \emph{Journal of The
  Royal Society Interface}, vol.~10, no.~85, Aug. 2013.

\bibitem{Olson:2013ko}
R.~S. Olson, D.~B. Knoester, and C.~Adami, ``{Critical interplay between
  density-dependent predation and evolution of the selfish herd},'' in
  \emph{Proceedings of the Genetic and Evolutionary Computation Conference
  (GECCO)}.\hskip 1em plus 0.5em minus 0.4em\relax ACM, Jul. 2013.

\bibitem{Fates:2013kp}
N.~Fat{\`e}s, ``{Stochastic Cellular Automata Solutions to the Density
  Classification Problem},'' \emph{Theory of Computing Systems}, vol.~53,
  no.~2, pp. 223--242, Aug. 2013.

\bibitem{Jin:2005cm}
Y.~Jin and J.~Branke, ``{Evolutionary Optimization in Uncertain Environments: A
  Survey},'' \emph{IEEE Transactions on Evolutionary Computation}, vol.~9,
  no.~3, pp. 303--317, Jun. 2005.

\bibitem{Capcarrere:2001kg}
M.~Capcarr{\`e}re and M.~Sipper, ``{Necessary conditions for density
  classification by cellular automata},'' \emph{Physical Review E}, vol.~64,
  no.~3, p. 036113, Aug. 2001.

\bibitem{Camazine:2003tb}
S.~Camazine, \emph{{Self-organization in Biological Systems - Google
  Books}}.\hskip 1em plus 0.5em minus 0.4em\relax Princeton University Press,
  2003.

\end{thebibliography}
\end{document}